\documentclass[preprint,12pt]{elsarticle}

\usepackage{amssymb}
\usepackage{amsmath}
\usepackage{booktabs}
\usepackage{tabularx}
\usepackage{array}
\usepackage{makecell}
\usepackage[table]{xcolor}
\usepackage{pifont}
\usepackage{adjustbox}
\usepackage{float}
\usepackage{xurl}
\usepackage{hyperref}

\newcommand{\present}{\cellcolor{gray!35}\ding{51}}
\newcommand{\detpresent}{\cellcolor{gray!35}\ding{51}}
\journal{Computers and Electronics in Agriculture}

\begin{document}

\begin{frontmatter}



\title{A Dataset-Centric Benchmark of Deep Learning Methods for Grape Leaf Disease Classification and Detection} 


\author[1]{Petar Canoski} 

\author[1]{Vlatko Spasev} 

\author[1]{Ivica Dimitrovski} 

\author[1]{Ivan Kitanovski} 

\author[1]{Petre Lameski} 

\affiliation[1]{organization={Faculty of Computer Science and Engineering, University Ss Cyril and Methodius},
            addressline={ul. Rudzer Boshkovikj 16, P.O. 393}, 
            city={Skopje},
            postcode={100}, 
            country={North Macedonia}}

\begin{abstract}
Grape leaf disease recognition is an important task in precision agriculture, supporting early diagnosis, timely intervention, and improved vineyard management. Deep learning methods have achieved strong results in plant disease analysis; however, many existing studies rely on a limited number of datasets, often acquired under controlled conditions. Consequently, reported performance may not fully reflect the challenges of real vineyard environments, where images are affected by complex backgrounds, variable illumination, occlusion, leaf pose, disease severity, and differences in acquisition devices. This paper presents a dataset-centric benchmark of deep learning methods for grape leaf disease classification and detection. We provide a structured analysis of publicly available datasets, considering their disease categories, annotation types, acquisition conditions, image characteristics, class distributions, provenance, and suitability for different recognition tasks. Based on this analysis, representative deep learning models are evaluated in three complementary settings: image-level classification, region-level classification, and object detection. Image-level and region-level classification are assessed using accuracy, while object detection is evaluated using mAP@50 and mAP@50:95. In addition to within-dataset evaluation, cross-dataset experiments examine whether models transfer between datasets with compatible disease categories but different visual and annotation characteristics. The results reveal near-saturated classification performance on several controlled or derivative datasets, greater difficulty on more heterogeneous datasets, and substantial variation in detection performance across annotation settings. Cross-dataset performance decreases sharply, particularly for object detection, showing that shared disease labels do not necessarily define equivalent recognition tasks. The benchmark highlights the importance of dataset provenance, realistic field evaluation, annotation compatibility, and external validation when developing reliable systems for vineyard disease classification and detection.
\end{abstract}


\end{frontmatter}

\section{Introduction}

Grapevine production is one of the most important sectors of modern horticulture and viticulture, contributing to fresh fruit production, wine production, rural economies, and agri-food value chains~\cite{santos2020review}. However, grapevine cultivation is highly vulnerable to biotic stresses, including fungal, bacterial, and viral diseases that can affect leaves, shoots, clusters, and the overall physiological condition of the plant. Leaf diseases are particularly important because visible symptoms on leaves often provide early indicators of infection and plant stress~\cite{bhargava2024plant},~\cite{portela2024systematic}. If not detected and managed in time, grapevine diseases can reduce yield, degrade fruit quality, increase production costs, and negatively affect the long-term sustainability of vineyard systems.

Early, accurate, and scalable disease detection is therefore essential for precision viticulture. In conventional vineyard management, disease diagnosis is usually performed through visual inspection by farmers, agronomists, or plant protection experts. Although expert inspection remains highly valuable, it is also time-consuming, subjective, labor-intensive, and difficult to apply consistently over large vineyard areas~\cite{davcev2026agentic}. The reliability of manual diagnosis can be affected by the experience of the observer, disease severity, symptom similarity between diseases, illumination conditions, occlusion, and the spatial variability of infection within the vineyard. These limitations motivate the development of automated image-based systems that can support decision-making, reduce monitoring costs, and enable more timely plant protection interventions~\cite{annabel2019machine}.

Recent advances in agricultural engineering, computer vision, machine learning, and sensing technologies have created new opportunities for automated crop monitoring. Digital cameras, mobile devices, unmanned aerial vehicles, ground robots, and edge-computing platforms are increasingly used to collect visual data in agricultural environments~\cite{spasev2023semantic}. In this context, deep learning methods have achieved strong performance in plant disease recognition, particularly for image-level classification tasks where a model assigns a disease label to an input image \cite{mohanty2016using, ferentinos2018deep, kamilaris2018deep}. For grapevine monitoring, deep learning models have been applied to identify diseases such as black rot, Esca, leaf blight, downy mildew, powdery mildew, and healthy leaf conditions. These developments show the potential of artificial intelligence to support precision viticulture and contribute to more efficient, data-driven vineyard management.

Despite these advances, the practical deployment of grape leaf disease recognition systems remains challenging. Many existing studies report high classification accuracy using datasets collected under controlled conditions, where leaves are photographed against simple backgrounds and under relatively stable illumination~\cite{santos2020review}. While such datasets are useful for initial model development and comparison, they do not fully represent the complexity of real vineyard environments. Field-acquired images often contain complex backgrounds, overlapping leaves, variable lighting, shadows, motion blur, different camera viewpoints, partial occlusions, mixed disease symptoms, and symptoms at different development stages. As a result, models trained only on controlled datasets may learn dataset-specific visual patterns rather than disease-specific features, leading to reduced generalization when applied to real vineyard images.

Another important limitation is that many grape leaf disease studies focus only on image-level classification. Classification is useful when an image contains a single dominant leaf or symptom, but it provides limited information in practical field scenarios where multiple leaves, disease symptoms, and background objects may appear in the same image~\cite{zhang2025lightweight}. Spatially annotated datasets can support different experimental settings depending on the meaning and completeness of their annotations. When annotations identify selected regions of interest but do not comprehensively cover all visible symptoms, the annotated regions may be more appropriately used as inputs for region-level classification. In this setting, localization is provided by the annotation and the model only predicts the category of the extracted region. In contrast, object detection requires the model to jointly predict the locations and categories of disease targets in the complete image.

Object detection methods are therefore more suitable for field scenarios containing multiple leaves or localized symptoms and can support vineyard monitoring systems, robotic inspection platforms, and decision-support tools~\cite{WISAENG2025113871}. However, spatial annotations such as bounding boxes and masks are more expensive and time-consuming to produce, and their suitability for detection depends on sufficiently clear target definitions and comprehensive annotation coverage. Consequently, publicly available datasets for grape leaf disease detection are less common than classification datasets, and their annotation schemes, disease categories, acquisition conditions, and target definitions are often not directly comparable.

From an agricultural engineering perspective, the performance of a deep learning model cannot be evaluated independently of the dataset used for training and testing. Dataset characteristics such as acquisition environment, image resolution, class distribution, disease categories, annotation granularity, label consistency, image provenance, and field variability can directly affect reported performance. A model that performs well on one controlled image-level classification dataset may not transfer reliably to an independently acquired field dataset. Similarly, a detector trained using one annotation protocol may perform poorly on another dataset even when the nominal disease categories overlap, because the datasets may differ in target scale, bounding-box definition, symptom appearance, and annotation density. Dataset overlap and derivative content can further make apparently independent evaluations less informative. Therefore, a dataset-centric evaluation is necessary to understand better the strengths, limitations, and generalization of deep learning methods for grape leaf disease recognition.

Benchmarking is also important for reproducibility and practical model selection~\cite{dimitrovski2023current},~\cite{dimitrovski2023aitlas}. In applied agricultural engineering research, it is not sufficient to report only the highest result obtained on a single dataset. A useful benchmark should compare representative models under a consistent experimental protocol, apply metrics appropriate to each task, and analyze model behaviour across datasets with different characteristics. In this study, image-level and region-level classification are evaluated using accuracy, while object detection is evaluated using mAP@50 and mAP@50:95. The two classification settings are reported separately because complete images and manually localized regions represent different input units. Similarly, detection results are interpreted separately for each dataset because their bounding boxes represent different semantic targets, including local symptom regions and complete diagnostically relevant leaves.

This paper presents a dataset-centric benchmark of deep learning methods for grape leaf disease classification and detection. The study first analyzes publicly available grape leaf disease image datasets according to their disease taxonomy, annotation type, acquisition conditions, image characteristics, class distribution, provenance, and suitability for different recognition tasks. Based on this analysis, representative deep learning models are evaluated under task-specific but internally consistent experimental protocols.

The benchmark distinguishes three complementary settings. Image-level classification assigns one disease category to a complete input image. Region-level classification assigns a category to a manually localized crop extracted from a spatial annotation. Object detection processes the complete image and jointly predicts the locations and categories of disease targets. In addition to within-dataset evaluation, cross-dataset experiments are conducted for compatible image-level classification datasets and for the shared downy- and powdery-mildew categories of the detection datasets. Dataset-overlap analysis is also performed to identify exact image reuse among controlled or derivative classification datasets.

The main objective of this study is not simply to rank model architectures, but to clarify how dataset properties influence evaluation outcomes, cross-dataset robustness, and the practical interpretation of grape leaf disease recognition performance. By jointly considering controlled and field-oriented acquisition, dataset provenance, class balance, disease-label compatibility, annotation granularity, annotation completeness, and domain shift, the paper aims to provide practical guidance for researchers developing computer vision systems for precision viticulture.

The main contributions of this paper are:
\begin{itemize}
    \item We provide a dataset-centric analysis of publicly available grape leaf disease image datasets, considering disease taxonomy, acquisition conditions, image characteristics, annotation granularity, class distribution, dataset provenance, and suitability for image-level classification, region-level classification, or object detection.

    \item We benchmark representative convolutional, transformer-based, and hybrid convolutional--transformer models for image-level grape leaf disease classification under a unified training and evaluation protocol, using accuracy as the primary evaluation metric.

    \item We introduce a separate region-level classification setting for spatially annotated datasets whose annotations identify selected regions of interest but do not provide sufficiently exhaustive coverage for conventional object-detection evaluation.

    \item We benchmark representative YOLO-family and transformer-based object detectors on multiple grape disease detection datasets, using mAP@50 and mAP@50:95 to evaluate detection and localization performance.

    \item We investigate cross-dataset generalization for both image-level classification and object detection after harmonizing compatible disease categories across datasets.

    \item We analyze how controlled versus field acquisition, exact image overlap, class imbalance, annotation semantics, target granularity, and dataset shift affect the interpretation of benchmark results.
\end{itemize}

The remainder of the paper is organized as follows. Section~\ref{sec:related_work} reviews related work on grape leaf disease classification, detection and localization, lightweight and transformer-based models, and cross-dataset generalization. Section~\ref{sec:datasets} analyzes the publicly available datasets and explains their use for image-level classification, region-level classification, or object detection. Section~\ref{sec:methodology} presents the problem formulation, dataset harmonization procedures, benchmark design, and evaluated model architectures. Section~\ref{sec:experimental_setup} describes the dataset partitions, training protocols, evaluation metrics, and implementation details. Section~\ref{sec:results} presents and discusses the image-level and region-level classification results, object-detection results, dataset-overlap analysis, cross-dataset experiments, and qualitative Gradient-weighted Class Activation Mapping (Grad-CAM) interpretation. Finally, Section~\ref{sec:conclusion} summarizes the main findings, limitations, and future research directions.

\section{Related Work}
\label{sec:related_work}
The terms \textit{detection}, \textit{identification}, \textit{diagnosis}, and \textit{recognition} are not used consistently in the grape disease literature. Several studies describe their methods as disease detection even though the final output is a single class label for an image or a previously segmented region. In this review, image-level classification refers to assigning one label to a complete image, region-level classification refers to classifying a manually localized crop, and object detection refers specifically to jointly predicting object locations and class labels in a complete image. The terminology used in the original studies is retained when describing their contributions, while their experimental task is clarified where necessary.

\subsection{Classical Image Processing and Machine Learning Approaches}

Early research on grape leaf disease recognition relied mainly on image processing, handcrafted feature extraction, and classical machine learning classifiers. These approaches typically followed a multi-stage pipeline consisting of leaf segmentation, diseased-region extraction, feature computation, feature selection, and final classification. Although such methods require more manual design than deep learning models, they remain important because they provide interpretable processing steps and can be applied when only small datasets are available.

Meunkaewjinda et al. proposed an early hybrid intelligent system for grape leaf disease recognition from color imagery \cite{meunkaewjinda2008grape}. The method combined grape leaf color segmentation, disease-region segmentation, self-organizing maps, genetic algorithms, Gabor wavelet features, and support vector machines (SVMs). The system classified grape leaf images into scab disease, rust disease, and no disease, demonstrating the feasibility of automated machine vision for agricultural inspection. A neural-network-based grape leaf disease diagnosis system was presented in \cite{sannakki2013diagnosis}. In that work, grape leaf images with complex backgrounds were processed using thresholding and anisotropic diffusion to remove noise. K-means clustering was then used to segment diseased regions, and a feed-forward back-propagation neural network was applied for classification. This study represents an early combination of image segmentation and neural classification for grape disease diagnosis.

Padol and Yadav proposed an SVM-based grape leaf disease recognition pipeline \cite{padol2016svm}. Their approach first used K-means clustering to identify diseased regions and then extracted color and texture features from the segmented areas. The final classification was performed using an SVM classifier, achieving a reported accuracy of 88.89\%. This study illustrates the dependence of classical methods on segmentation quality and handcrafted feature design. Waghmare et al. introduced a decision-support system for grape disease recognition based on opposite-colour local binary pattern features and multiclass SVM classification \cite{waghmare2016detection}. The method focused on downy mildew and black rot and used texture patterns extracted from segmented diseased regions. The study is relevant because it considered grape diseases that are important in practical vineyard management but are not always present in common PlantVillage-style benchmarks~\cite{hughes2015open}.

Jaisakthi et al. developed a machine-learning-based system for grape leaf disease identification using GrabCut segmentation, diseased-region extraction, and classifiers such as SVM, AdaBoost, and random forest \cite{jaisakthi2019grape}. The system classified leaves into healthy, rot, Esca, and leaf blight categories. Among the tested classifiers, SVM achieved the best reported testing accuracy of 93\%. An automatic K-means clustering and machine learning approach was proposed in \cite{javidan2023diagnosis}. In that study, diseased regions were separated from healthy leaf tissue using K-means clustering. Features were extracted from RGB, HSV, and L*a*b color spaces, and SVM classification was combined with principal component analysis (PCA)-based dimensionality reduction and Relief feature selection. The method was evaluated on black measles, black rot, and leaf blight and achieved high reported accuracy.

These classical image processing and machine learning studies demonstrate that grape leaf diseases can be recognized from color, texture, and segmented symptom regions. However, their performance depends strongly on image acquisition conditions, pre-processing quality, segmentation accuracy, and manually selected features. These limitations motivated the transition toward convolutional neural network (CNN)-based and transfer learning approaches.

\subsection{CNN and Transfer Learning Methods for Grape Leaf Disease Classification}

The development of large public plant disease datasets and deep convolutional neural networks significantly changed image-based plant disease recognition. Mohanty et al. demonstrated the feasibility of deep learning for plant disease diagnosis using the PlantVillage dataset \cite{mohanty2016using}. Their work trained CNN models on a large collection of healthy and diseased plant leaf images acquired under controlled conditions and reported very high within-dataset classification accuracy. This study became a major reference point for later plant disease classification research, including many grape leaf disease studies that use PlantVillage-derived datasets.

Most recent grape leaf disease studies formulate the task as image-level classification. In this setting, each image is assigned to one disease class or to a healthy class. A common experimental configuration uses the grape subset of PlantVillage, which includes healthy leaves, black rot, Esca or black measles, and leaf blight. This four-class setting has become a widely used baseline for CNN and transfer-learning models. Huang et al. developed modified deep learning models for grape leaf disease detection and classification using transfer learning with VGG16, MobileNet, and AlexNet \cite{huang2020grape}. Their study targeted black rot, black measles, leaf blight, and phylloxera. In addition to individual models, the authors proposed an ensemble model to improve classification performance. This work is relevant because it extended the standard PlantVillage-style classification setting by including phylloxera as an additional category.

Prasad et al. proposed a multiclass grape leaf disease classification method based on a deep CNN classifier using VGG16 with additional convolutional layers \cite{prasad2024multiclass}. The model was evaluated using accuracy and F1-score and achieved high classification performance. The study positioned the approach as a decision-support system for farmers, but the evaluation remained focused on image-level classification. Darmawan et al. introduced a hybrid transfer-learning and machine-learning approach based on EfficientNetB0 feature extraction and Categorical Boosting classification \cite{darmawan2025grape}. Instead of using EfficientNetB0 as an end-to-end classifier, the model used it to extract deep features, while Categorical Boosting was applied as the final classifier. This type of hybrid approach is relevant because it combines deep representation learning with classical machine learning classification.

The study in \cite{kunduracioglu2024advancements} evaluated a large number of CNN and vision transformer models for grape leaf disease classification and grapevine variety recognition. The authors used the PlantVillage grape subset for disease classification and a separate Grapevine dataset for variety recognition, including the Ak, Alaidris, Buzgulu, Dimnit, and Nazli grapevine varieties. This study is useful for demonstrating the strong performance of modern pre-trained architectures, but it also highlights that high accuracy is often obtained under dataset-specific conditions. A CNN and improved K-nearest neighbor-based grape leaf image classification approach was presented in \cite{shantkumari2023grape}. The study used PlantVillage images and extracted high-quality histogram and gradient-based features before classification. The method compared CNN-based classification and improved KNN against traditional classification models, showing that both learned and handcrafted features can contribute to grape leaf disease recognition.

DeepLeaf was proposed as an optimized deep learning approach for automated grapevine leaf disease recognition \cite{talaat2025deepleaf}. The method combined preprocessing, feature extraction, and an optimized CNN classifier whose hyperparameters were adjusted using fuzzy optimization. The study focused on leaf blight, black rot, black measles, and a healthy or stable class using PlantVillage images. A ResNet50V2-based grape plant disease classification method was proposed in \cite{kosaraju2026grape}. The model was trained to classify grape leaves into rot, Esca, healthy, and leaf blight categories and was compared with CNN and VGG16 baselines. The study emphasized the ability of residual learning to extract subtle local and global disease patterns from leaf images. These CNN and transfer-learning studies show that deep models can achieve high classification accuracy on grape leaf disease datasets. However, most evaluations are conducted within a single dataset, which limits conclusions about generalization to different acquisition conditions, external datasets, or field-acquired vineyard images.

\subsection{Lightweight, Attention-Based, Transformer-Based, and Hybrid Classification Models}
Recent work has increasingly focused on lightweight models, attention mechanisms, transformer-based architectures, and hybrid convolutional--transformer designs. These approaches broaden the range of representations available for grape disease classification, from efficient local feature extraction to global and hierarchical attention mechanisms. These methods aim to improve the balance between accuracy and computational efficiency, which is important for mobile, embedded, and edge-based agricultural applications. A MobileNetV2-based grape leaf disease classification method was proposed in \cite{sahid2025image}. The study classified grape leaves into healthy, black rot, Esca or black measles, and leaf blight categories. The authors emphasized computational efficiency and reported high classification performance, suggesting that lightweight CNNs are suitable for mobile and edge deployment. Another MobileNetV2-based real-time grape leaf disease classification approach was presented in \cite{mathew2025real}. This study considered black rot, black measles, Isariopsis leaf spot, and healthy leaves. The evaluation reported strong performance in terms of accuracy, precision, recall, specificity, F1-score, and receiver operating characteristic (ROC)-based analysis, supporting the use of lightweight CNNs for scalable disease monitoring.

A fine-grained grape disease recognition method based on GC-MobileNet was proposed in \cite{canghai2025fine}. The method modified MobileNetV3 using Ghost modules, convolutional block attention module (CBAM) attention, LeakyReLU activation, transfer learning, and data augmentation. The model was designed to reduce the number of parameters while improving robustness for disease types and severity levels, making it relevant for resource-constrained applications. An EfficientNetV2L-based grape leaf disease diagnosis method with data augmentation and gradient-weighted class activation mapping (Grad-CAM) visualization was introduced in \cite{venkatachalam2025advanced}. The study considered downy mildew, powdery mildew, black rot, and anthracnose. The use of Grad-CAM is relevant because it provides visual interpretability by showing which image regions influence the model prediction.

SwinGNet combined Swin Transformer features with GoogLeNet classification for real-time grape leaf disease recognition \cite{babu2025swingnet}. The model was deployed on a Raspberry Pi 5 GPU platform, making the study particularly relevant from an agricultural engineering perspective because it connects model design with practical edge deployment. A comparative optimization study evaluated DenseNet121, VGG19, VGG16, InceptionV3, and ResNet50V2 for grape leaf disease identification \cite{patil2025comparative}. DenseNet121 achieved the best reported performance among the tested architectures. Such comparative studies are useful because they highlight the importance of backbone selection, but they are still usually limited by the dataset used for evaluation. These studies demonstrate the expansion of grape disease classification from standard CNNs toward lightweight, attention-based, transformer-based, and hybrid architectures. Nevertheless, many evaluations remain model-centric and are conducted on a single dataset. Consequently, it is often unclear whether reported improvements result primarily from architectural design or from the characteristics of the selected dataset, and external validation under different acquisition conditions remains limited.

\subsection{Grape Leaf Disease Detection and Localization}

Image-level classification is useful when an image contains a single dominant leaf or symptom, but it does not provide spatial information about the location of the disease. For vineyard monitoring, object detection and localization are more practical because field images may contain multiple leaves, overlapping vegetation, complex backgrounds, shadows, and localized symptoms.

Spatial annotations do not necessarily imply that conventional object detection is the most appropriate experimental formulation. Bounding boxes or masks may comprehensively identify all instances in an image, in which case they can support quantitative object-detection evaluation. In other datasets, annotations may indicate only selected diagnostically relevant regions. Such annotations can instead be used to extract samples for region-level classification, where localization is supplied in advance and the model is evaluated only on its ability to classify the annotated region. The suitability of a spatially annotated dataset therefore depends on the semantics and completeness of its annotations, not only on the annotation format.

A real-time grape leaf disease detector based on improved convolutional neural networks was proposed by Xie et al. \cite{xie2020deep}. The authors constructed the grape leaf disease dataset (GLDD) and introduced Faster DR-IACNN, a Faster R-CNN-based detector enhanced with Inception-v1, Inception-ResNet-v2, and squeeze-and-excitation blocks. The detector addressed black rot, black measles, leaf blight, and mites, achieving 81.1\% mAP and 15.01 FPS.

Praveen et al. investigated grape leaf disease detection based on attention mechanisms \cite{praveen2023novel}. The study evaluated detectors such as Single Shot Multibox Detection (SSD), Faster R-CNN, and You Only Look Once (YOLO) with attention mechanisms including CBAM, squeeze-and-excitation networks, and efficient channel attention. The results showed that attention mechanisms can improve feature selection and detection performance. Zhang et al. proposed YOLOv5-CA for automatic grape downy mildew detection under field conditions \cite{zhang2022deep}. The method integrated coordinate attention into YOLOv5 to emphasize disease-related features. This study is important because it focused on downy mildew detection in natural vineyard environments and reported a favorable trade-off between detection accuracy and speed.

Grape Guard was proposed as a YOLO-based mobile application for grape leaf disease detection \cite{mamun2025grape}. The authors trained YOLOv5 and YOLOv8 models and converted the best-performing model to TensorFlow Lite for Android deployment. This work is relevant because it connects model training with an end-user mobile application for vineyard disease diagnosis. A YOLOv8-based method for detecting grape leaf black rot spots was introduced in \cite{zhu2025application}. The method incorporated Spatial Pyramid Dilated Convolution (SPD-Conv) and an efficient multi-scale attention module to improve detection of small target spots. The study evaluated the model using PlantVillage images and orchard images, showing the importance of testing under different image sources.

GCS-YOLO was proposed as a lightweight grape leaf disease detection algorithm based on improved YOLOv8 \cite{hu2025gcs}. The model introduced a lightweight C2f-GR feature extraction module, RepConv, CBAM attention, and an optimized detection head. The aim was to reduce parameters and computational load while maintaining or improving detection accuracy. YOLOv8-ACCW was introduced as another lightweight improved YOLOv8 model for grape leaf disease detection \cite{chen2024yolov8}. The method used AKConv, coordinate attention, Content-aware Reassembly of Features (CARAFE), and Wise-IoU (Weighted Interpolation of Sequential Evidence for Intersection over Union) loss to improve small lesion detection and mobile-device deployment. The targeted categories included black root, black measles, and blight.

Rossi et al. introduced LDD, a grape disease dataset for object detection and instance segmentation \cite{rossi2022ldd}. The dataset contains images of leaves and grape clusters with and without disease symptoms in natural vineyard context. It is particularly important for this paper because it provides instance-level annotations and enables evaluation beyond image-level classification. The HERMOS dataset was introduced as a spatially annotated resource for the visual analysis of grape leaf diseases \cite{ozacar2024hermos}. It contains field-acquired images with Pascal VOC bounding boxes identifying selected healthy regions and regions associated with dead arm, downy mildew, and powdery mildew~\cite{everingham2015pascal}. HERMOS is important because it
provides one of the few publicly available grape disease datasets with localized region annotations. However, as with any spatially annotated dataset, its suitability for object detection depends on whether the annotations define consistent targets and provide sufficiently comprehensive coverage. The final experimental use of this dataset is therefore examined in Section~\ref{sec:datasets}.

Ghiani et al. developed an automated system for detecting downy mildew and powdery mildew symptoms in grapevines \cite{ghiani2025automated}. Their work is relevant not only because it addresses mildew detection, but also because it explicitly emphasizes the role of data partitioning and dataset diversity in reliable model performance. Detection and localization studies are highly relevant for agricultural engineering applications because they provide spatial information that can support robotic scouting, mobile inspection, variable-rate treatment, and decision-support systems. However, their results are particularly sensitive to annotation quality and target definition. Some datasets annotate complete diseased leaves, whereas others annotate local spots, lesions, symptom clusters, or selected regions of interest. Reported mAP values may also be based on different data partitions, intersection-over-union thresholds, and annotation conversions. Consequently, the presence of spatial annotations does not make detection results directly comparable across datasets, and selectively annotated datasets may be more appropriately evaluated through region-level classification.

\subsection{Dataset Bias, Disease-Label Heterogeneity, and Cross-Dataset Generalization}

Dataset bias is a central limitation in grape leaf disease recognition. Many studies report high accuracy using controlled or semi-controlled datasets, but such results may not transfer to field-acquired vineyard images. Models trained on controlled images may learn background appearance, illumination patterns, image resolution, acquisition device characteristics, or dataset-specific artifacts instead of disease-specific symptoms.

The PlantVillage dataset has played an important role in plant disease recognition research, but it also illustrates the limitations of controlled datasets~\cite{hughes2015open}. Mohanty et al. showed that deep learning models can achieve very high accuracy on PlantVillage under controlled evaluation conditions \cite{mohanty2016using}. However, the same work also indicated that performance decreases when models are tested on images collected under different conditions. This supports the need for more diverse and field-representative evaluation.

PlantDoc was introduced to address the lack of realistic non-lab plant disease datasets \cite{singh2020plantdoc}. Unlike controlled datasets, PlantDoc contains plant disease images collected under more natural conditions with non-trivial backgrounds. Although it is not grape-specific, it is relevant to this paper because it demonstrates the importance of evaluating disease recognition models under realistic image acquisition conditions. Noyan specifically investigated bias in the PlantVillage dataset \cite{noyan2022uncovering}. The study showed that a model trained using only a small number of background pixels could achieve accuracy far above random guessing, indicating that label-correlated background or acquisition bias exists in the dataset. This finding is directly relevant to grape disease classification studies that rely heavily on PlantVillage-style images.

Dataset provenance represents an additional but less frequently examined source of bias. Public datasets may be repackaged, augmented, renamed, or derived from earlier collections, and the same original image may therefore appear in multiple nominally different datasets. Exact or near-duplicate images shared between training and evaluation data can lead to information leakage and produce overly optimistic estimates of model generalization~\cite{barz2020we,kapoor2023leakage}. Dataset-centric evaluation should therefore consider not only visual-domain differences and class-label compatibility, but also source provenance, derivative relationships, and image overlap.

Ahmad et al. studied the generalization of deep learning-based plant disease identification under controlled and field conditions \cite{ahmad2023toward}. Their work evaluated models across multiple corn disease datasets, including controlled and field-acquired data. Although the study is not grape-specific, it provides important evidence that cross-dataset evaluation is necessary for assessing whether plant disease models are suitable for deployment.

Gui et al. investigated automatic field plant disease recognition and addressed the effect of field image variability on model performance \cite{gui2021towards}. Their study is relevant because it considered practical problems such as background complexity and proposed strategies to improve recognition in field environments. This supports the broader argument that dataset characteristics must be considered when evaluating plant disease recognition systems.

Disease-label heterogeneity further complicates dataset-centric evaluation. The reviewed grape leaf disease classification studies often use the PlantVillage-style four-class configuration consisting of healthy, black rot, Esca or black measles, and leaf blight. This disease set appears in several CNN and transfer-learning studies, including MobileNetV2-based classification \cite{sahid2025image}, ResNet50V2-based classification \cite{kosaraju2026grape}, and DeepLeaf \cite{talaat2025deepleaf}.

Other studies use different disease categories. Huang et al. considered black rot, black measles, leaf blight, and phylloxera \cite{huang2020grape}. The EfficientNetV2L study considered downy mildew, powdery mildew, black rot, and anthracnose \cite{venkatachalam2025advanced}. The Nashik grape leaf disease dataset included healthy leaves, downy mildew, powdery mildew, and bacterial leaf spot \cite{dharrao2025grapes}. The real-time MobileNetV2 study included black rot, black measles, Isariopsis leaf spot, and healthy leaves \cite{mathew2025real}.

Detection studies also differ in disease and target definitions. Xie et al. considered black rot, black measles, leaf blight, and mites \cite{xie2020deep}. Zhang et al. focused specifically on grape downy mildew detection \cite{zhang2022deep}. The YOLOv8 black rot study focused on black rot spots \cite{zhu2025application}. YOLOv8-ACCW addressed black root, black measles, and blight \cite{chen2024yolov8}. LDD includes multiple grape disease-related object categories for detection and instance segmentation \cite{rossi2022ldd}, while HERMOS includes powdery mildew, downy mildew, dead arm or dead root, and healthy leaves \cite{ozacar2024hermos}.

This heterogeneity makes cross-dataset evaluation difficult. Esca and black measles may be treated as equivalent in some datasets but as different labels in others. Leaf blight and Isariopsis leaf spot may also be used inconsistently. Mildew-related diseases such as downy mildew and powdery mildew are agronomically important but are not always included in common benchmark datasets. Pest-related categories such as mites and phylloxera further complicate label harmonization because they do not always correspond to the same type of leaf disease label.

Therefore, cross-dataset generalization should be considered an essential evaluation component for grape leaf disease recognition. Models trained on controlled datasets should be evaluated on independently acquired field-oriented data whenever compatible categories are available. Similarly, transfer between field datasets is needed to assess robustness to changes in illumination, background, camera device, cultivar, disease severity, leaf pose, and symptom presentation.

For object detection, compatible disease names alone do not guarantee a compatible evaluation task. Datasets may use different annotation units, target scales, bounding-box construction rules, annotation densities, or criteria for deciding which visible symptoms should be annotated. A detector may therefore fail to transfer even when the source and target datasets use the same disease labels. Cross-dataset detection should consequently be interpreted as transfer across both visual domains and annotation protocols.

Overall, the reviewed literature demonstrates substantial progress in grape leaf disease recognition, including classical image-processing pipelines, CNN and transfer-learning classifiers, lightweight and attention-based networks, transformer-based and hybrid architectures, interpretable classification methods, instance-segmentation datasets, and modern object detectors. However, most studies remain focused on a single model family, a single dataset, or one annotation setting. Reported results are therefore difficult to compare because of differences in disease taxonomy, acquisition conditions, dataset provenance, class distribution, annotation granularity, target definition, data partitioning, and evaluation metrics.

Relatively few studies jointly examine image-level classification, classification of manually localized regions, and complete-image object detection while also considering dataset overlap and cross-dataset transfer. This motivates the dataset-centric benchmark proposed in this paper, where datasets are treated not only as sources of training and evaluation samples, but also as experimental factors that determine task formulation, comparability, and generalization.

\section{Public Datasets for Grape Leaf Disease Recognition}
\label{sec:datasets}
The proposed benchmark is built around publicly available grape leaf disease datasets with different annotation granularities and experimental uses. Dataset properties strongly influence reported model performance; therefore, the datasets are analyzed not only as sources of training and evaluation images, but also as experimental factors affecting robustness, comparability, and task formulation. In particular, we consider differences in disease taxonomy, annotation unit, annotation completeness, acquisition conditions, pre-processing, predefined splits, class balance, image variability, dataset provenance, and possible image overlap.

Three evaluation settings are distinguished. Image-level classification assigns one disease category to a complete input image. Region-level classification assigns a category to a manually localized crop extracted from a spatial annotation. Object detection processes the complete image and jointly predicts the locations and categories of visible disease targets. The presence of bounding boxes or masks alone does not determine which task is appropriate; the decision also depends on the semantic meaning, consistency, and coverage of the annotations. The following subsections describe the image-level classification datasets, the spatially annotated datasets and their experimental use, and the principal dataset-related challenges relevant to benchmark design.

\subsection{Image-Level Classification Datasets}
\label{subsec:classification_datasets}
Table~\ref{tab:classification_datasets} summarizes the publicly available datasets considered for grape leaf disease classification. These datasets differ substantially in acquisition conditions, image dimensions, pre-processing procedures, label taxonomy, and dataset scale. Such variation is particularly important in the context of dataset-centric benchmarking, because the reported performance of deep learning models is influenced not only by model architecture, but also by the visual properties and annotation design of the underlying dataset. In this subsection, classification refers specifically to image-level classification, where each complete input image is associated with a single class label. Datasets constructed from manually annotated regions or cropped
patches are considered separately because their input unit and localization assumptions differ from those of complete-image classification.

\begin{table}[H]
\centering
\caption{Characteristics of the image-level classification datasets used in the benchmark. \textit{Name} identifies the original dataset or grape-specific subset; \textit{\#Images} reports the total number of images included in the processed benchmark version; \textit{\#Classes} gives the number of retained image-level categories; \textit{Image size} specifies the native or released image dimensions, or indicates that dimensions vary across images; \textit{Image condition} summarizes the acquisition or pre-processing setting, including field acquisition, controlled backgrounds, background removal, or data augmentation; and \textit{Predefined splits} indicates whether the dataset provides an official training, validation, and test partition.}
\label{tab:classification_datasets}
\small
\begin{adjustbox}{max width=\linewidth}
\begin{tabular}{
    >{\raggedright\arraybackslash}p{4.0cm}
    c
    c
    c
    >{\raggedright\arraybackslash}p{3.3cm}
    c
}
\toprule
\textbf{Name} & \textbf{\#Images} & \textbf{\#Classes} & \textbf{Image size} & \textbf{Image condition} & \textbf{Predefined splits}\\
\midrule
GVLiD~\cite{gayakwad2026curated} & 3,477 & 4 & 1,080~$\times$~1,080 & Field & No \\
NGLD~\cite{dharrao2025grapes} & 2,726 & 4 & 256~$\times$~256 & Field/mobile; background removed & No \\
PlantVillage~\cite{mohanty2016using} & 4,062 & 4 & 256~$\times$~256 & Controlled & No \\
PDR2018~\cite{aichallenger2018crop} & 3,167 & 4 & varying & Controlled & Yes \\
GLDD~\cite{truprojects2023grape} & 1,598 & 4 & 256~$\times$~256 & Controlled & Yes \\
GLDD Augmented~\cite{kavcmar2025augmented} & 18,054 & 4 & 600~$\times$~800 & Augmented & Yes \\
GLHCD~\cite{anan2024grape} & 21,007 & 5 & varying & Controlled; augmented & No \\
New Plant Diseases~\cite{bhattarai2018newplant} & 9,027 & 4 & 256~$\times$~256 & Controlled; augmented & Yes \\
GL-Portugal~\cite{portela2026dataset} & 5,267 & 5 & 1,024~$\times$~1,024 & Field & No \\
\bottomrule
\end{tabular}
\end{adjustbox}
\end{table}

A first distinction can be made between controlled datasets, field-acquired datasets, and preprocessed or augmented variants. The PlantVillage (grape subset)~\cite{mohanty2016using}, the New Plant Diseases Dataset (grape subset)~\cite{bhattarai2018newplant}, the Grape Leaf Disease Dataset (GLDD)~\cite{truprojects2023grape}, and parts of the Grape Leaf Health Classification Dataset (GLHCD)~\cite{anan2024grape} largely reflect controlled or visually simplified acquisition settings. NGLD~\cite{dharrao2025grapes} was originally captured using mobile phones, but the released images were resized and processed using background removal to obtain a uniform background and emphasize the leaf region. Therefore, although NGLD originates from mobile image acquisition, its released form is closer to a field-derived but visually standardized classification dataset. In contrast, GVLiD~\cite{gayakwad2026curated} provides field-acquired vineyard images with richer field variability and associated metadata, making it particularly relevant for evaluating robustness under realistic vineyard conditions. Similarly, the grapevine leaf image dataset from Portugal (GL-Portugal)~\cite{portela2026dataset} provides high-resolution RGB images captured under natural lighting in experimental and commercial vineyards in northern Portugal. Its field-oriented acquisition protocol preserves natural variability in illumination, leaf orientation, canopy structure, and background conditions, making it relevant for evaluating classification under realistic vineyard environments.

GVLiD~\cite{gayakwad2026curated} and GL-Portugal~\cite{portela2026dataset} provide the most direct representation of natural vineyard conditions, whereas NGLD~\cite{dharrao2025grapes} is field-derived but visually standardized through resizing and background removal. GVLiD is a curated vineyard dataset containing 3,477 high-resolution grapevine leaf images captured directly under vineyard field conditions, with expert-verified annotations and accompanying metadata, including environmental and acquisition information. NGLD~\cite{dharrao2025grapes} contains 2,726 images of table grape leaves across four classes: healthy leaves, downy mildew, powdery mildew, and bacterial leaf spot. Although the images were originally captured using mobile phones, the released dataset applies resizing and background removal to produce a more uniform visual appearance. This pre-processing reduces background noise and emphasizes the leaf region, making the dataset suitable for image classification, but also reducing some of the natural background variability typically present in field-acquired images. GL-Portugal~\cite{portela2026dataset} contains 5,267 high-resolution RGB images acquired directly from four experimental and commercial vineyard locations in northern Portugal. The dataset includes healthy leaves and leaves exhibiting downy mildew, powdery mildew, Esca complex, and erineum mite symptoms. The images were captured using a smartphone camera under natural daytime illumination during the 2024 and 2025 growing seasons. Unlike datasets subjected to background removal or synthetic background replacement, GL-Portugal retains natural vineyard backgrounds and variations in illumination, leaf orientation, and surrounding vegetation.

Another important distinction concerns the disease taxonomy. Several datasets, including the GVLiD, PlantVillage, the New Plant Diseases Dataset, PDR2018, GLDD, and the Grape Leaf Disease Augmented Dataset (GLDD Augmented), follow the common four-class configuration of healthy leaves, black rot, Esca or black measles, and leaf blight or isariopsis leaf spot. This shared structure makes them useful for within-group benchmarking and for cross-dataset experiments after harmonizing label names. For this purpose, disease names are harmonized into canonical categories. In PDR2018, severity variants such as ``general'' and ``serious'' are grouped into their corresponding disease categories. Esca and black measles are treated as a harmonized category, while leaf blight and Isariopsis leaf spot are grouped because these labels are often used interchangeably in grape leaf disease datasets.

However, not all datasets follow this taxonomy. NGLD includes healthy leaves, downy mildew, powdery mildew, and bacterial leaf spot, while GLHCD contains healthy leaves, brown spot, mites, black rot, and downy mildew. GL-Portugal partially overlaps with the common taxonomy through its healthy and Esca complex categories, while extending the disease coverage with downy mildew, powdery mildew, and erineum mite symptoms. These alternative label sets broaden the represented disease spectrum but limit direct cross-dataset comparison to compatible class subsets. Table~\ref{tab:classification_disease_coverage} summarizes the disease-category coverage of the classification datasets. For visualization, dataset-specific mite categories, including erineum-mite symptoms, are grouped under the broader category of mite-related symptoms. It shows that most datasets follow the common four-class configuration of healthy leaves, black rot, Esca or black measles, and leaf blight or Isariopsis leaf spot, whereas NGLD, GLHCD, and GL-Portugal introduce additional categories such as downy mildew, powdery mildew, bacterial leaf spot, brown spot, and mite-related symptoms.

\begin{table}[H]
\centering
\caption{Disease-category coverage across grape leaf disease classification datasets.
Filled cells indicate that a disease category is included in the dataset.}
\label{tab:classification_disease_coverage}

\small
\setlength{\tabcolsep}{2.3pt}
\renewcommand{\arraystretch}{1.15}

\begin{adjustbox}{max width=\linewidth}
\begin{tabular}{
    >{\raggedright\arraybackslash}p{3.6cm}
    >{\centering\arraybackslash}m{1.15cm}
    >{\centering\arraybackslash}m{1.15cm}
    >{\centering\arraybackslash}m{1.30cm}
    >{\centering\arraybackslash}m{1.35cm}
    >{\centering\arraybackslash}m{1.15cm}
    >{\centering\arraybackslash}m{1.20cm}
    >{\centering\arraybackslash}m{1.30cm}
    >{\centering\arraybackslash}m{1.35cm}
    >{\centering\arraybackslash}m{1.15cm}
}

\toprule

\textbf{Dataset} &
{\scriptsize\textbf{Healthy}} &
{\scriptsize\textbf{Black rot}} &
{\scriptsize\textbf{Esca / Black measles}} &
{\scriptsize\textbf{Leaf blight / Isariopsis}} &
{\scriptsize\textbf{Downy mildew}} &
{\scriptsize\textbf{Powdery mildew}} &
{\scriptsize\textbf{Bacterial leaf spot}} &
{\scriptsize\textbf{Mite-related symptoms}} &
{\scriptsize\textbf{Brown spot}} \\

\midrule

GVLiD \cite{gayakwad2026curated} & \present & \present & \present & \present & & & & & \\
NGLD \cite{dharrao2025grapes} & \present & & & & \present & \present & \present & & \\
PlantVillage \cite{mohanty2016using} & \present & \present & \present & \present & & & & & \\
PDR2018 \cite{aichallenger2018crop} & \present & \present & \present & \present & & & & & \\
GLDD \cite{truprojects2023grape} & \present & \present & \present & \present & & & & & \\
GLDD Augmented \cite{kavcmar2025augmented} & \present & \present & \present & \present & & & & & \\
GLHCD \cite{anan2024grape} & \present & \present & & & \present & & & \present & \present \\
New Plant Diseases \cite{bhattarai2018newplant} & \present & \present & \present & \present & & & & & \\
GL-Portugal \cite{portela2026dataset} & \present & & \present & & \present & \present & & \present & \\
\bottomrule
\end{tabular}
\end{adjustbox}

\end{table}

The datasets also vary substantially in scale. With 5,267 field-acquired images, GL-Portugal represents a medium-sized collection that is larger than GVLiD and the smaller controlled datasets, while its size is based on images acquired under natural vineyard conditions rather than synthetic augmentation. Smaller datasets such as the Grape Leaf Disease Dataset provide a relatively limited number of images, whereas larger collections such as the New Plant Diseases Dataset and the Grape Leaf Disease Augmented Dataset provide many more training examples. However, dataset size should be interpreted carefully. In several cases, larger datasets are obtained through augmentation or synthetic generation rather than through independent image acquisition. As a result, increased image counts do not necessarily imply increased visual diversity. This distinction is important when interpreting benchmark results, especially in cases where models may benefit from repeated or highly similar augmented samples.

In terms of image characteristics, most classification datasets use relatively small image dimensions, frequently around $256 \times 256$ pixels, while GVLiD and GL-Portugal provide substantially higher-resolution images. GL-Portugal is distributed at $3,000 \times 3,000$ pixels and in a reduced $1,024 \times 1,024$ version, preserving substantially more visual detail than the smaller controlled datasets. This difference may influence the visibility of fine-grained symptoms and the effectiveness of pre-processing and resizing pipelines. Predefined train/validation/test splits are also inconsistent across datasets. Some datasets provide official splits, while others do not, requiring the benchmark to define reproducible partitioning strategies. This inconsistency further motivates a unified evaluation protocol across all considered datasets.

Dataset provenance also requires explicit consideration. Some classification collections are augmented, repackaged, or derived from previously released datasets, and nominally different dataset names do not necessarily indicate independent image content. Exact decoded-image hashing was therefore used to audit image overlap among the classification datasets. This analysis does not alter the within-dataset evaluation protocol, but it is used when interpreting cross-dataset results because shared images can make transfer performance appear stronger than evaluation on an independent target domain. The overlap findings are reported together with the cross-dataset classification results.

\begin{figure}[H]
\centering
\includegraphics[
    width=\linewidth,
    height=0.95\textheight,
    keepaspectratio
]{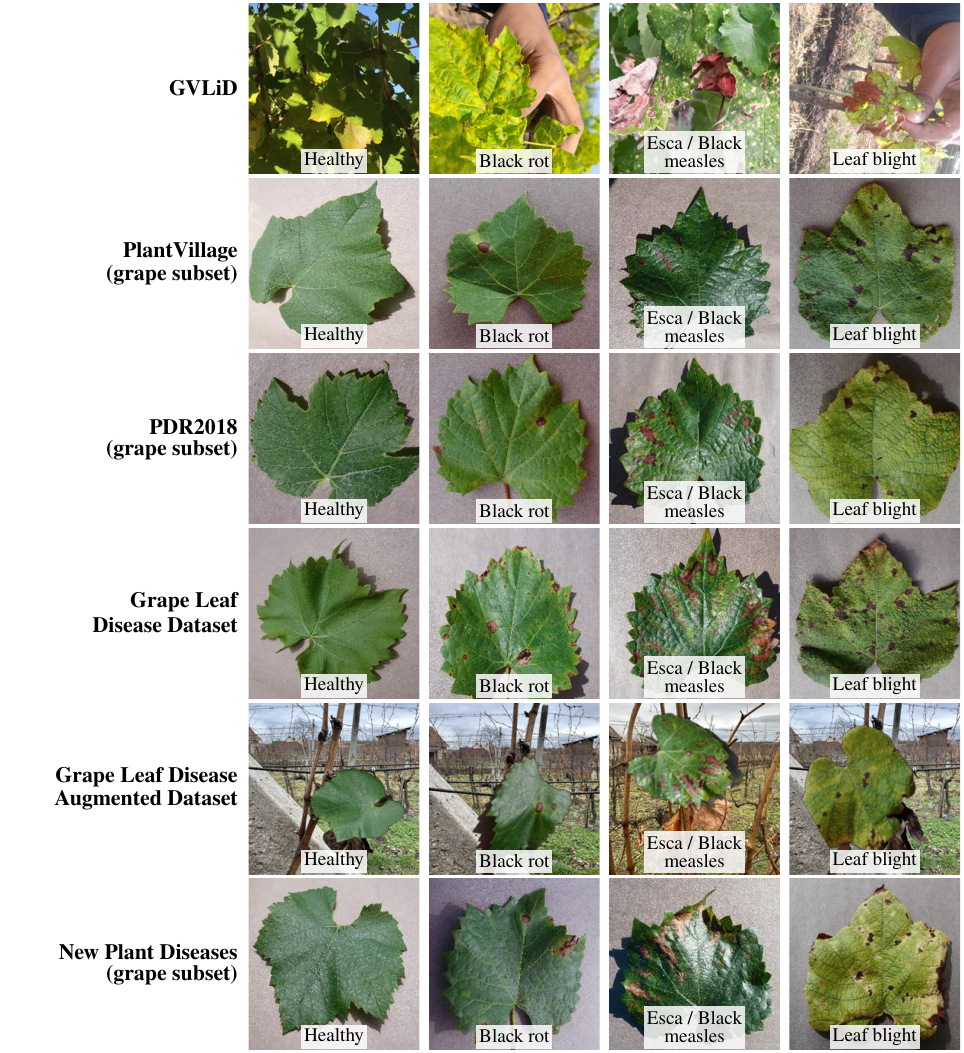}
\caption{Representative images from grape leaf disease classification datasets following the common four-class taxonomy. Each row corresponds to one dataset, while the columns show healthy leaves, black rot, Esca or black measles, and leaf blight or Isariopsis leaf spot.}
\label{fig:classification_examples_common}
\end{figure}

\begin{figure}[H]
\centering
\includegraphics[
    width=\linewidth,
    height=0.95\textheight,
    keepaspectratio
]{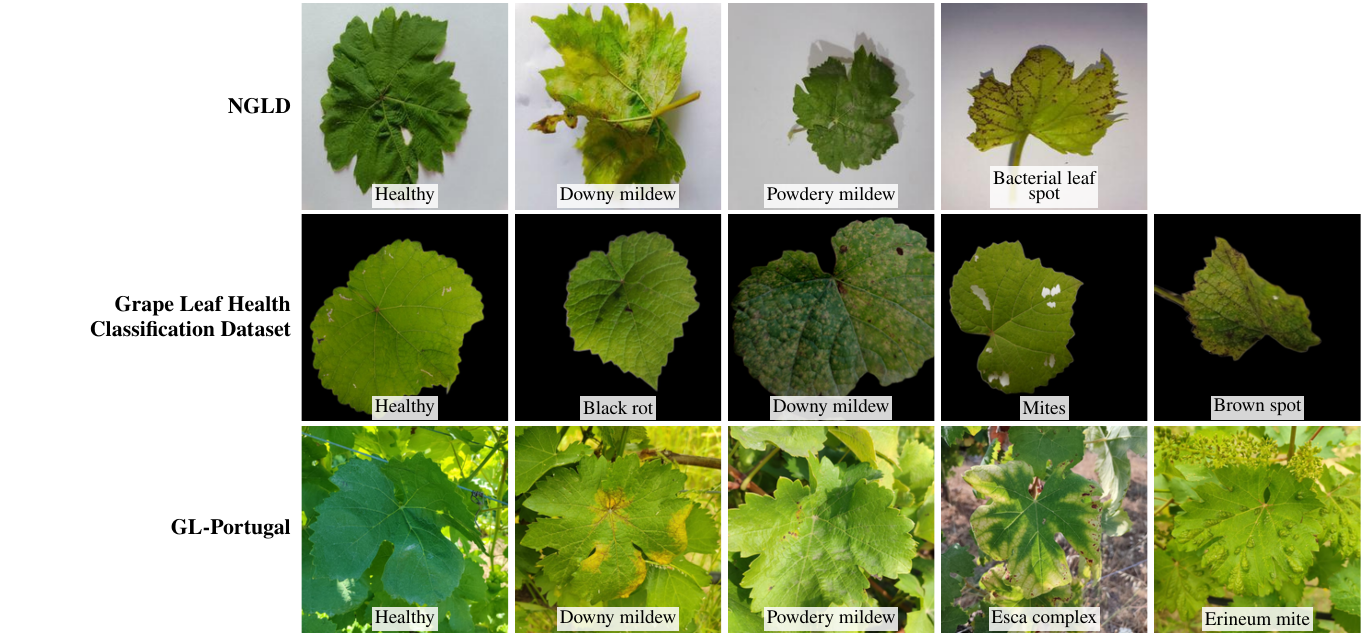}
\caption{Representative images from classification datasets with alternative grape leaf disease taxonomies. NGLD contains healthy leaves, downy mildew, powdery mildew, and bacterial leaf spot; GLHCD contains healthy leaves, black rot, downy mildew, mites, and brown spot; and GL-Portugal contains healthy leaves, downy mildew, powdery mildew, Esca complex, and erineum mite symptoms.}
\label{fig:classification_examples_alternative}
\end{figure}

Figures~\ref{fig:classification_examples_common} and
\ref{fig:classification_examples_alternative} present representative images from the classification datasets. Figure~\ref{fig:classification_examples_common} focuses on datasets following the common four-class taxonomy, whereas Figure~\ref{fig:classification_examples_alternative} shows datasets with broader or partially overlapping disease categories. Together, the figures highlight substantial differences in scene complexity, background conditions, color distribution, pre-processing, and symptom appearance. In particular, the contrast between controlled or visually standardized datasets and field-acquired vineyard datasets illustrates the need for cross-dataset evaluation. The examples shown in these figures correspond to the complete images used as model inputs. They should therefore be distinguished from region-level classification samples, which are extracted from spatial annotations before being provided to the classifier.

\begin{figure}[H]
\centering
\includegraphics[width=\textwidth]
{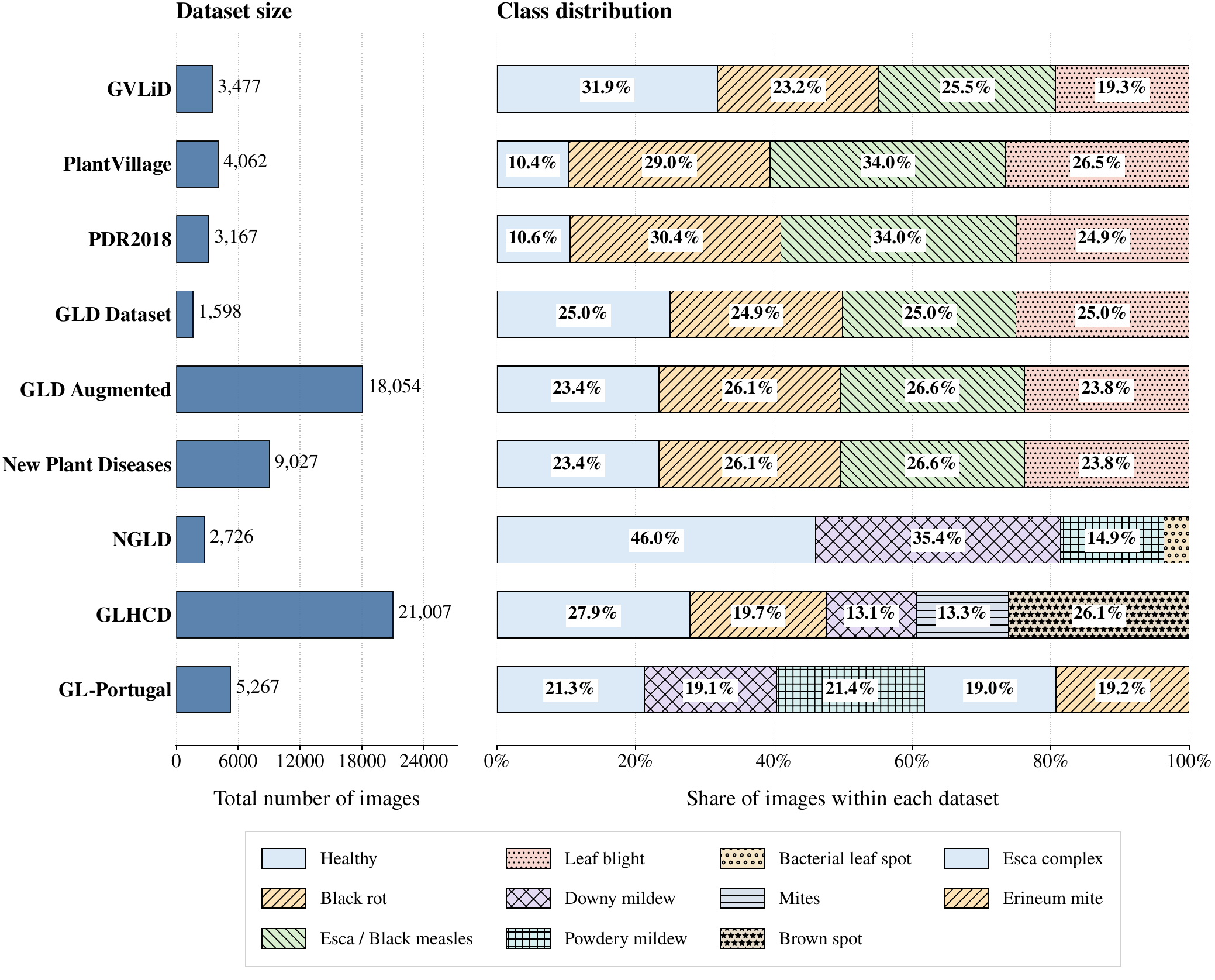}
\caption{Dataset size and class distribution of the grape leaf disease classification datasets included in the benchmark. The left panel reports the total number of images in each dataset, aggregated across all available partitions, while the right panel shows the percentage of images assigned to each class. Colors and hatch patterns identify the disease categories. The figure highlights differences in dataset scale, class balance, and disease-category coverage across datasets with common, alternative, or partially overlapping taxonomies.}
\label{fig:classification_distributions}
\end{figure}

Figure~\ref{fig:classification_distributions} complements the representative image examples by summarizing both dataset scale and class distribution across all image-level classification datasets. The left panel reports the total number of images in each dataset, while the right panel shows the relative class distribution within each dataset. The results reveal substantial variation in both dataset size and class balance. The Grape Leaf Disease Dataset is almost perfectly balanced, with each class representing approximately 25\% of the images. The Grape Leaf Disease Augmented Dataset and the New Plant Diseases grape subset are also relatively balanced, with class proportions between 23.4\% and 26.6\%. GL-Portugal similarly exhibits a balanced five-class distribution, with individual categories accounting for approximately 19.0\%--21.4\% of the images.

In contrast, NGLD is strongly imbalanced: healthy leaves and downy mildew represent 46.0\% and 35.4\% of the dataset, respectively, whereas bacterial leaf spot accounts for only 3.7\%. PlantVillage and PDR2018 also contain substantially fewer healthy images, which represent approximately 10\% of each dataset. Figure~\ref{fig:classification_distributions} also highlights the considerable differences in dataset scale, ranging from fewer than 2,000 images in the smallest processed dataset to approximately 20,000 images in the largest collections. These differences are important because both class imbalance and dataset scale can influence model training, class-specific recognition performance, and the interpretation of aggregate evaluation results.

Overall, the classification datasets considered in this study provide a diverse but fragmented resource landscape. They collectively cover both standard four-class grape disease recognition and broader disease settings, but they also exhibit substantial heterogeneity in acquisition condition, disease taxonomy, scale, and split availability. These differences make the classification benchmark scientifically meaningful, while also reinforcing the need for careful dataset harmonization and transparent experimental design. A separate region-level classification setting is introduced in Section~\ref{subsubsec:region_level_datasets}. Unlike the datasets discussed above, that setting uses annotated image regions as classification inputs and is therefore analyzed independently from the image-level benchmark.

\subsection{Spatially Annotated Datasets and Benchmark Use}
\label{subsec:spatially_annotated_datasets}
Spatially annotated datasets provide bounding boxes, polygon masks, or other region-level annotations that identify diagnostically relevant leaves or local disease symptoms. However, spatial annotation availability does not automatically imply that a dataset is suitable for conventional object-detection evaluation. Object detection requires sufficiently clear target definitions and sufficiently comprehensive annotation coverage so that predicted instances can be reliably matched with reference instances. When annotations instead identify selected regions of interest, the same spatial information may be more appropriately used to extract crops for region-level classification. The datasets considered in this study are therefore organized according to their final experimental use rather than annotation format alone.

\subsubsection{Region-Level Classification Datasets}
\label{subsubsec:region_level_datasets}
Region-level classification represents an intermediate setting between image-level classification and object detection. Instead of assigning a single label to a complete image, the classifier receives a spatially localized region extracted from an annotated source image. The localization is therefore provided in advance, and the learning task is limited to predicting the category of the selected region. This formulation is particularly suitable for datasets in which annotations identify representative healthy or symptomatic areas but do not necessarily provide comprehensive coverage of every visible target in the complete image.

Among the publicly available datasets retained in this benchmark, HERMOS~\cite{ozacar2024hermos} is used for region-level classification. It contains field-acquired grapevine images with Pascal VOC bounding-box annotations for selected healthy and symptomatic regions associated with dead arm, downy mildew, and powdery mildew. The annotations describe local regions of interest rather than complete image-level labels. A single source image may therefore contribute multiple classification samples belonging to one or more region categories. After annotation verification and region extraction, the processed dataset contained 490 source images, from which 13,852 annotated regions were obtained across four classes. This corresponds to an average of approximately 28.3 annotated regions per source image. The extracted regions were used as the input samples for the region-level classification experiment.

The region distribution is not necessarily balanced. The processed dataset contains 5,060 healthy regions (approximately 36\%), 1,882 dead-arm regions (14\%), 1,392 downy mildew regions (10\%), and 5,518 powdery mildew regions (40\%). Differences in the number of annotated regions per class may influence training behaviour and class-specific recognition performance. In addition, region counts should be distinguished from source image counts because several regions may originate from the same photograph.

\begin{figure}[H]
\centering
\centering
\includegraphics[
    width=0.7\linewidth,
    keepaspectratio
]{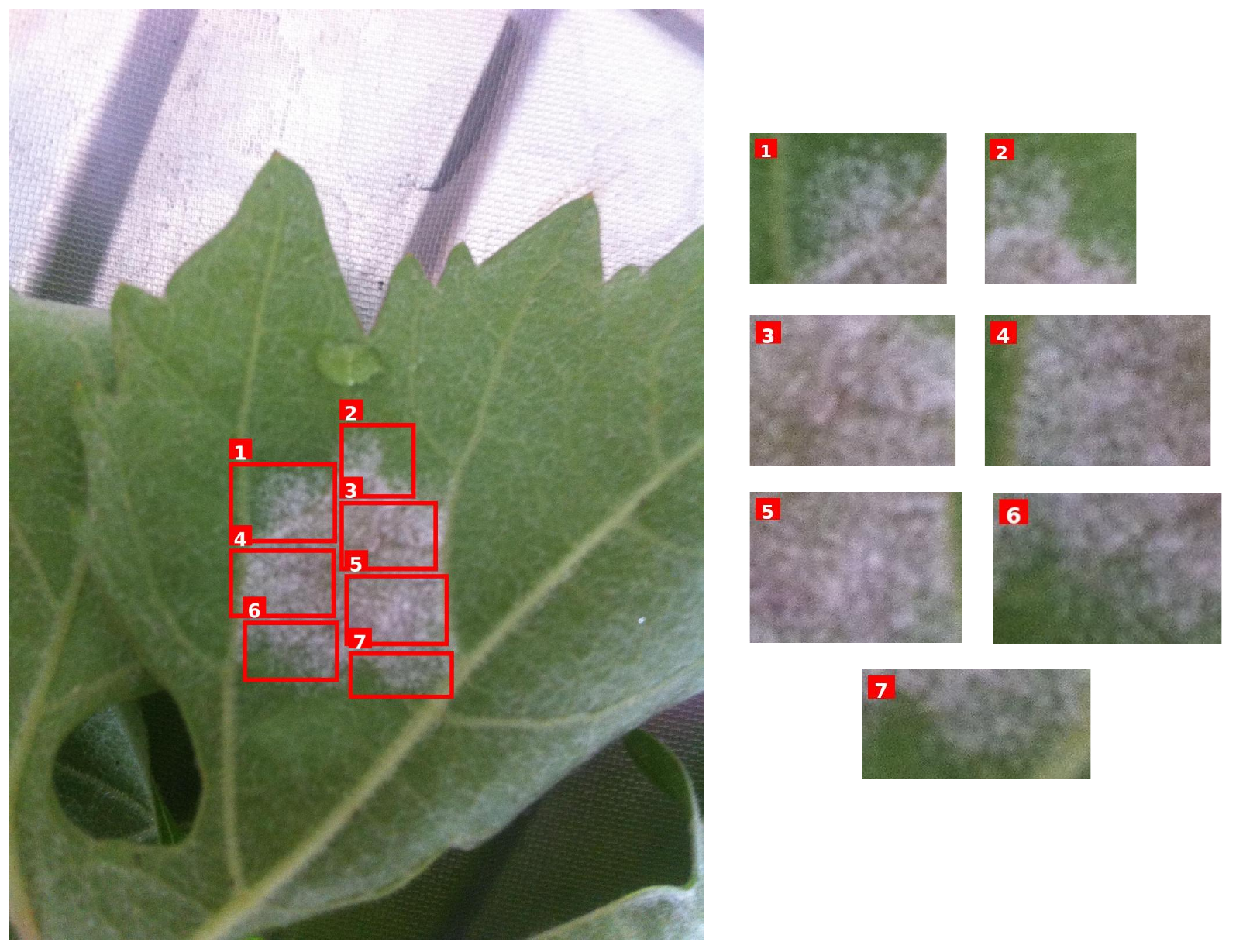}
\caption{Representative example from the HERMOS region-level classification dataset. The left panel shows a field-acquired source image with numbered annotated regions, while the right panel presents the corresponding extracted patches used as classification inputs. Matching numbers link each bounding box to its extracted region. In this example, the selected annotations correspond to powdery-mildew regions.}
\label{fig:region_classification_examples}
\end{figure}

Figure~\ref{fig:region_classification_examples} presents a representative source image, its annotated regions, and the corresponding extracted classification samples. The examples illustrate variation in region size, symptom appearance, illumination, background content, and the amount of contextual leaf tissue included within the bounding boxes.

Manual inspection showed that the HERMOS annotations do not comprehensively cover every visible symptomatic region in the source images. This limits their suitability for conventional object-detection evaluation, since predictions on visible but unannotated symptoms could be counted as false positives. Region-level classification does not require exhaustive annotation coverage because the model is evaluated only on explicitly annotated and extracted regions. The resulting performance is therefore reported separately from both the image-level classification and object-detection benchmarks. Region-level accuracy measures discrimination among already localized regions and should not be interpreted as directly comparable with classification accuracy on complete images or with object-detection mAP.

\subsubsection{Object Detection Benchmark Datasets}
\label{subsubsec:object_detection_datasets}

Object detection requires spatial annotations that allow a model to jointly predict the categories and locations of disease manifestations in complete images. This setting is particularly relevant to field monitoring, where several leaves or symptoms may occur within the same scene. However, the three datasets retained for the quantitative benchmark differ substantially in their annotation units. The Mildew symptom dataset~\cite{ghiani2025automated} uses directly defined bounding boxes around local symptomatic regions, the filtered LDD subset~\cite{rossi2022ldd} uses bounding boxes derived from polygon annotations of local leaf symptoms, and FD-Confounders~\cite{tardif2023expertized} annotates complete diagnostically relevant leaves within whole-vine scenes. Consequently, their results are interpreted according to the annotation definition of each dataset rather than as directly interchangeable detection tasks. Table~\ref{tab:detection_datasets} summarizes the final benchmark versions after dataset-specific filtering, annotation conversion, duplicate checking, and split construction.

\begin{table}[H]
\centering
\caption{Summary of the grape leaf disease datasets used in the object detection benchmark.
\textit{Dataset} identifies the original resource or the processed subset used in this study;
\textit{\#Images} reports the number of unique images in the final benchmark version;
\textit{\#Classes} gives the number of retained detection categories;
\textit{\#Annotations} reports the total number of retained bounding-box targets;
\textit{Image dimensions} describes the native image size or indicates that dimensions vary across images;
\textit{Annotation unit} specifies the semantic target represented by each bounding box, such as a local symptom region or a complete diagnostically relevant leaf; and
\textit{Split protocol} describes how the training, validation, and test partitions were defined.
Counts refer to the processed benchmark subsets after annotation conversion, category filtering, and duplicate checking.}
\label{tab:detection_datasets}

\small
\setlength{\tabcolsep}{2.5pt}
\renewcommand{\arraystretch}{1.15}

\begin{adjustbox}{max width=\linewidth}
\begin{tabular}{
    >{\raggedright\arraybackslash}p{3.1cm}
    >{\centering\arraybackslash}m{1.35cm}
    >{\centering\arraybackslash}m{1.35cm}
    >{\centering\arraybackslash}m{1.65cm}
    >{\centering\arraybackslash}m{2.0cm}
    >{\raggedright\arraybackslash}p{3.0cm}
    >{\raggedright\arraybackslash}p{4.1cm}
}
\toprule

\textbf{Dataset} &
{\scriptsize\textbf{\#Images}} &
{\scriptsize\textbf{\#Classes}} &
\makecell[c]{\scriptsize\textbf{\#Annota-}\\[-1pt]\scriptsize\textbf{tions}} &
\makecell[c]{\scriptsize\textbf{Image}\\[-1pt]\scriptsize\textbf{dimensions}} &
\makecell[l]{\scriptsize\textbf{Annotation}\\[-1pt]\scriptsize\textbf{unit}} &
{\scriptsize\textbf{Split protocol}} \\

\midrule

Mildew symptom dataset \cite{ghiani2025automated} & 1,404 & 2 & 24,775 & Varying & Local disease-symptom regions & Experiment-based partitions provided by the authors \\
LDD leaf-only subset \cite{rossi2022ldd} & 716 & 4 & 9,231 & Varying & Leaf symptom regions; polygon-derived boxes & Original validation retained as test; original training divided into training and validation \\
FD-Confounders \cite{tardif2023expertized} & 632 & 3 & 28,038 & $2,448 \times 2,048$ & Complete diagnostically relevant leaves & Grouped multilabel-stratified 70/15/15 split \\
\bottomrule
\end{tabular}
\end{adjustbox}

\vspace{1mm}

\parbox{\linewidth}{%
\footnotesize
\textit{Note:} Annotations belonging to organs or categories outside the leaf-detection scope were excluded where necessary.
}

\end{table}

The Mildew symptom dataset~\cite{ghiani2025automated} contains images of grapevines affected by downy mildew and powdery mildew, acquired between May and July at different stages of plant development. The downy mildew images were collected primarily from a field collection containing approximately 50 grapevine varieties, whereas the powdery mildew subset represents only three varieties. Images were captured from distances ranging from approximately 0.5~m to slightly more than 1~m and include both field-grown and potted plants. The potted plants were photographed mainly outdoors under varying illumination conditions and were positioned close together to reproduce the dense foliage typically encountered in vineyards. Some plants were also cultivated in large containers, allowing their growth structure to more closely resemble that of vines planted directly in the field.

The benchmark uses the 2022 subset, comprising 1,404 images with local bounding-box annotations for downy mildew and powdery mildew symptoms. The annotations correspond to visible symptomatic regions rather than complete leaves. The final converted version contains 24,775 retained annotations across the two disease classes. Because individual images may contain multiple small, spatially distributed symptom regions, the dataset represents a dense symptom-level detection task. The experiment-based partitions provided with the dataset were preserved to reduce the possibility that images acquired from closely related plants, varieties, or acquisition conditions were distributed across the training and evaluation subsets.

\begin{figure}[H]
\centering
\includegraphics[width=\textwidth]
{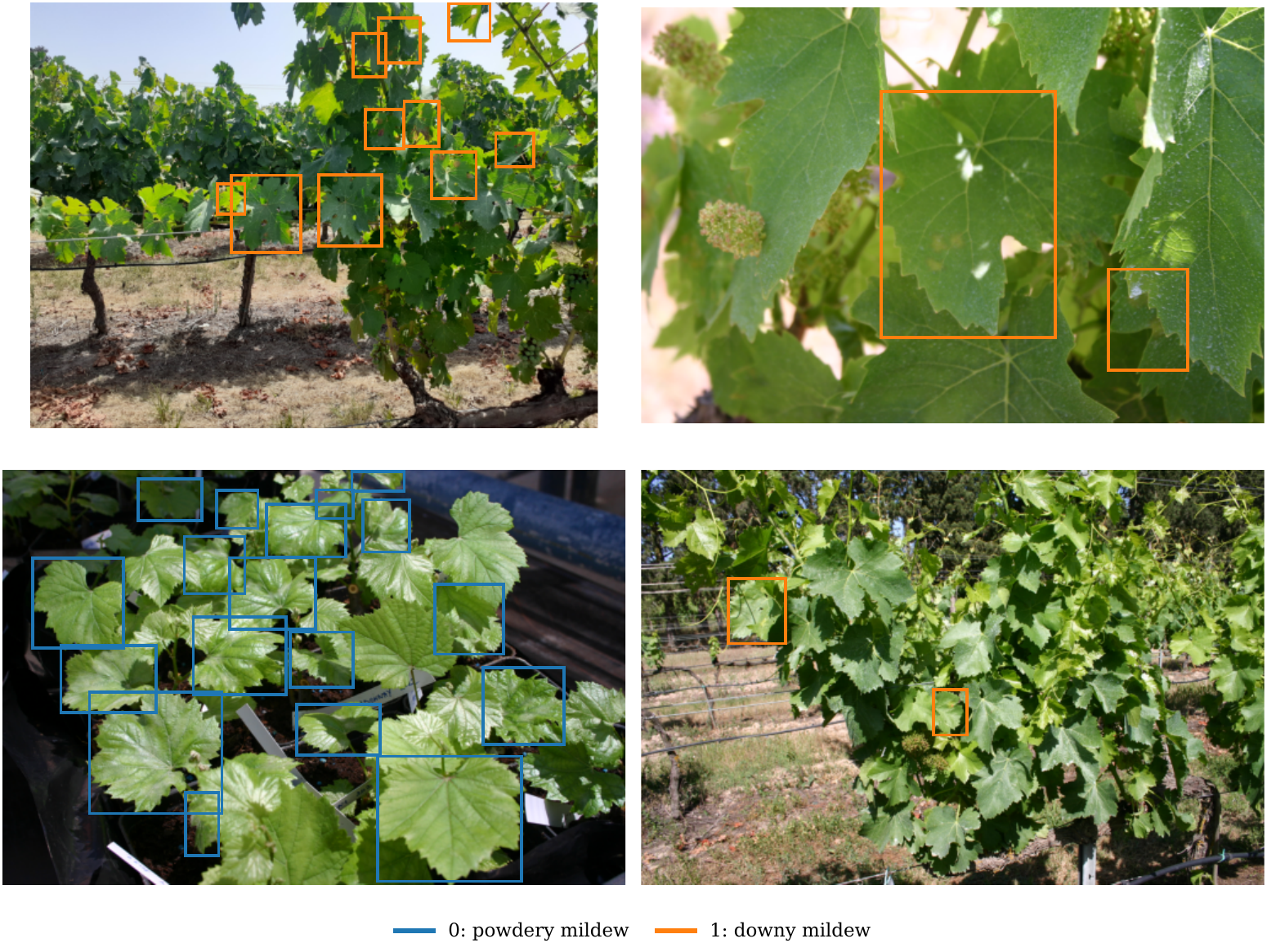}
\caption{Representative annotated examples from the Mildew symptom dataset. The images include field-grown and outdoor potted grapevines photographed under varying illumination and foliage-density conditions. Bounding boxes indicate local downy mildew and powdery mildew symptoms.}
\label{fig:mildew_detection_examples}
\end{figure}

Figure~\ref{fig:mildew_detection_examples} presents four representative annotated images from the Mildew symptom dataset. The examples illustrate variation in plant growth stage, acquisition distance, illumination, foliage density, and the size and spatial distribution of the annotated mildew symptoms.

LDD~\cite{rossi2022ldd} was originally developed for object detection and instance segmentation of grapevine organs and diseases. The dataset provides polygon masks together with their enclosing bounding boxes. Its annotations operate at two levels. Generic organ categories delineate complete leaves and grape bunches, whereas disease-specific categories delineate visible symptomatic regions occurring on these organs. Consequently, the complete dataset combines whole-organ localization with local symptom-region localization.

To align LDD with the leaf-disease scope of the present benchmark, a strict subset was constructed using only the leaf-specific disease categories: black rot, grey mould, powdery mildew, and downy mildew. Images containing grape-bunch or shoot categories were excluded, while generic complete-leaf annotations were removed. The retained polygons therefore correspond to local symptomatic regions on leaves rather than to complete leaf instances. For the common object-detection protocol, the enclosing bounding box of each retained polygon was used as the detection target.

The resulting LDD subset contains 716 images and 9,231 leaf-disease annotations across four classes. The supplied validation subset was retained as the final test set, while the filtered original training subset was divided into training and validation partitions using image-level multilabel stratification. This procedure preserves the original evaluation separation while providing a dedicated validation subset for model selection. Grey mould is substantially less represented than the other classes, and its per-class average precision should therefore be interpreted cautiously.

\begin{figure}[H]
\centering
\includegraphics[width=\textwidth]
{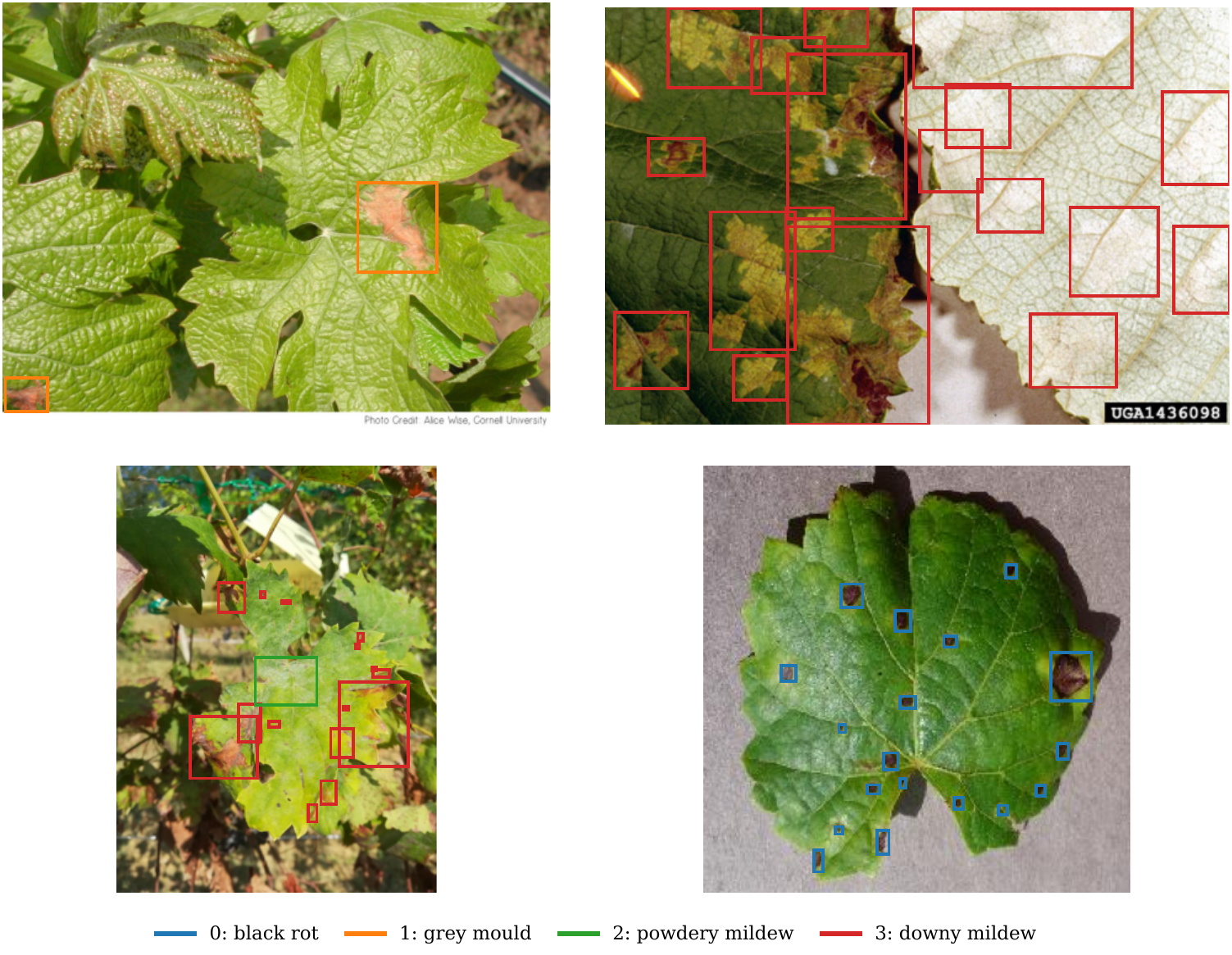}
\caption{Representative annotated examples from the filtered LDD leaf-disease subset. Only black rot, grey mould, powdery mildew, and downy mildew annotations on leaves were retained.}
\label{fig:ldd_detection_examples}
\end{figure}

Figure~\ref{fig:ldd_detection_examples} shows representative examples from the filtered LDD subset. Similar to the Mildew symptom dataset, the retained LDD annotations represent local disease manifestations rather than complete diseased leaves. However, the two datasets differ in annotation design and visual composition. LDD provides polygon-level delineations from which bounding boxes are derived, and its symptom regions vary substantially in scale, ranging from small isolated lesions to larger affected leaf areas. The images also exhibit considerable variation in resolution, background complexity, and symptom density.

FD-Confounders~\cite{tardif2023expertized} originates from a broader expert-annotated collection of whole-vine RGB images focused on Flavescence dorée and visually similar diseases, pests, and abiotic stresses. For the present benchmark, the \textit{soft annotation} subset was used because it provides bounding boxes around diagnostically relevant leaves. The original labels were harmonized into three classes: FD-symptomatic leaf, Esca-symptomatic leaf, and confounding leaf. Annotations describing symptomatic bunches and the heterogeneous category ``other'' were excluded because they do not correspond to the leaf-level detection task.

The FD-Confounders subset used in this study contains 632 unique images and 28,038 bounding-box annotations across three classes: FD-symptomatic leaves, Esca-symptomatic leaves, and leaves showing confounding symptoms. Because no official split is provided for this formulation, the dataset was divided into training, validation, and test sets using a grouped 70/15/15 partition that preserved the class distribution and kept duplicate or closely related images within the same subset.

\begin{figure}[t]
\centering
\includegraphics[width=\textwidth]
{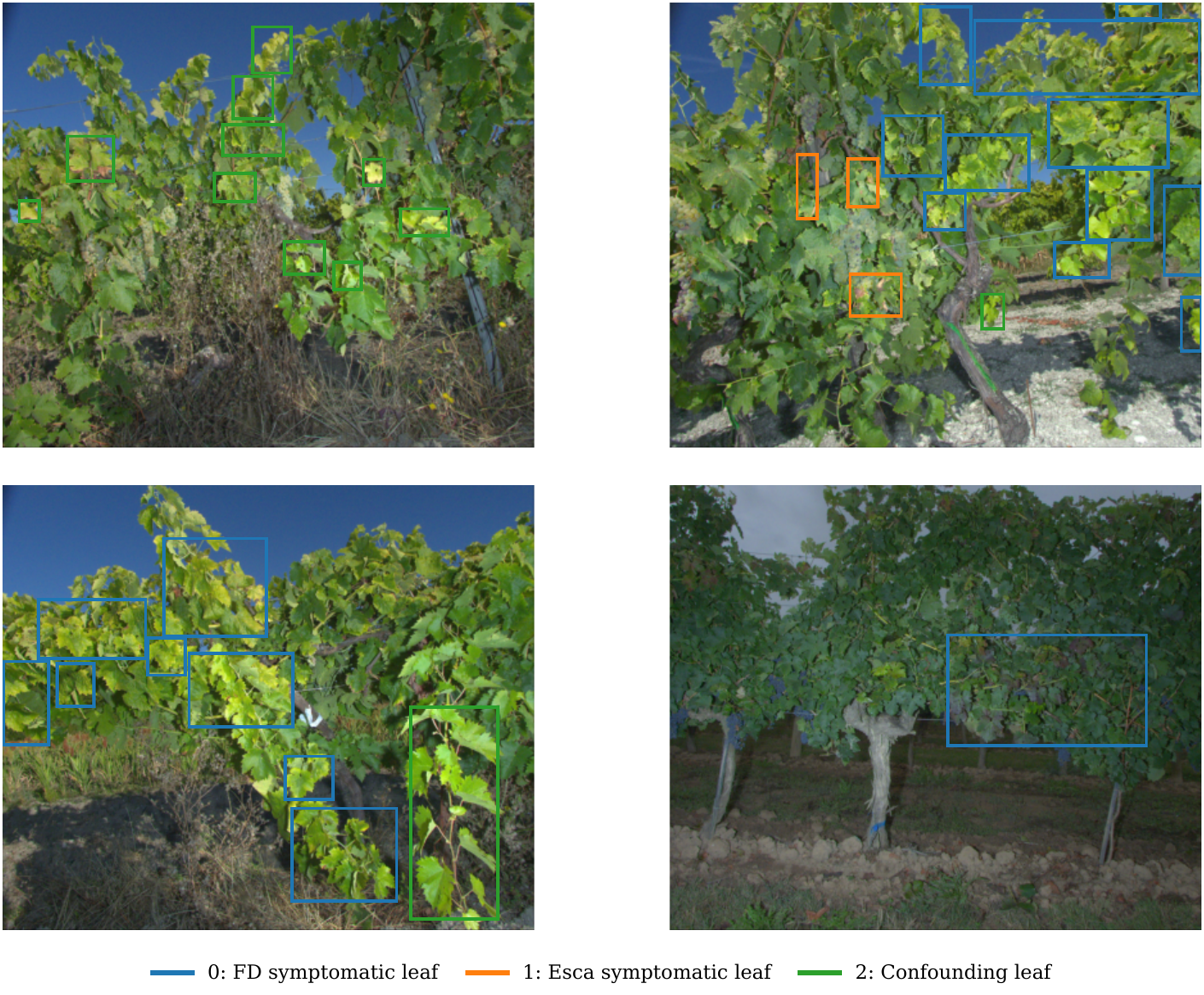}
\caption{Representative annotated examples from FD-Confounders. Bounding boxes correspond to FD-symptomatic leaves, Esca-symptomatic leaves, and leaves displaying confounding symptoms within whole-vine scenes.}
\label{fig:fd_confounders_detection_examples}
\end{figure}

Figure~\ref{fig:fd_confounders_detection_examples} illustrates the whole-vine acquisition conditions of FD-Confounders. Unlike the Mildew symptom dataset and LDD, which annotate local symptomatic regions, this dataset primarily localizes complete diagnostically relevant leaves. The presence of numerous leaves within complex vineyard scenes makes the task both a localization problem and a fine-grained distinction between FD-related, Esca-related, and confounding symptoms.

Table~\ref{tab:detection_disease_coverage} summarizes the target-category coverage of the three detection datasets. Unlike the classification benchmark, healthy tissue is not represented as an object category and is instead treated as background. The table also demonstrates that the three datasets have only limited class overlap. The Mildew symptom dataset and LDD share downy mildew and powdery mildew, while FD-Confounders uses a distinct diagnostic taxonomy centred on Flavescence dorée, Esca, and visually confounding symptoms.

\begin{table}[t]
\centering
\caption{Retained target-category coverage of the object detection datasets.
Filled cells indicate that the corresponding object category is included in the processed benchmark version.}
\label{tab:detection_disease_coverage}

\small
\setlength{\tabcolsep}{2.5pt}
\renewcommand{\arraystretch}{1.15}

\begin{adjustbox}{max width=\linewidth}
\begin{tabular}{
    >{\raggedright\arraybackslash}p{3.4cm}
    >{\centering\arraybackslash}m{1.25cm}
    >{\centering\arraybackslash}m{1.30cm}
    >{\centering\arraybackslash}m{1.35cm}
    >{\centering\arraybackslash}m{1.45cm}
    >{\centering\arraybackslash}m{1.55cm}
    >{\centering\arraybackslash}m{1.55cm}
    >{\centering\arraybackslash}m{1.55cm}
}
\toprule

\textbf{Dataset} &
\makecell[c]{\scriptsize\textbf{Black}\\[-1pt]\scriptsize\textbf{rot}} &
\makecell[c]{\scriptsize\textbf{Grey}\\[-1pt]\scriptsize\textbf{mould}} &
\makecell[c]{\scriptsize\textbf{Downy}\\[-1pt]\scriptsize\textbf{mildew}} &
\makecell[c]{\scriptsize\textbf{Powdery}\\[-1pt]\scriptsize\textbf{mildew}} &
\makecell[c]{\scriptsize\textbf{FD-}\\[-1pt]
             \scriptsize\textbf{symptomatic}\\[-1pt]
             \scriptsize\textbf{leaf}} &
\makecell[c]{\scriptsize\textbf{Esca-}\\[-1pt]
             \scriptsize\textbf{symptomatic}\\[-1pt]
             \scriptsize\textbf{leaf}} &
\makecell[c]{\scriptsize\textbf{Confounding}\\[-1pt]
             \scriptsize\textbf{leaf}} \\

\midrule

Mildew symptom dataset & & & \detpresent & \detpresent & & & \\
LDD leaf-only subset & \detpresent & \detpresent & \detpresent & \detpresent & & & \\
FD-Confounders & & & & & \detpresent & \detpresent & \detpresent \\
\bottomrule
\end{tabular}
\end{adjustbox}

\vspace{1mm}

\parbox{\linewidth}{%
\footnotesize
\textit{Note:} Healthy tissue is treated as background rather than as a detectable object.
The confounding category in FD-Confounders represents a heterogeneous group of symptoms
that may visually resemble Flavescence dorée.
}

\end{table}

\begin{figure}[H]
\centering
\includegraphics[width=\textwidth]
{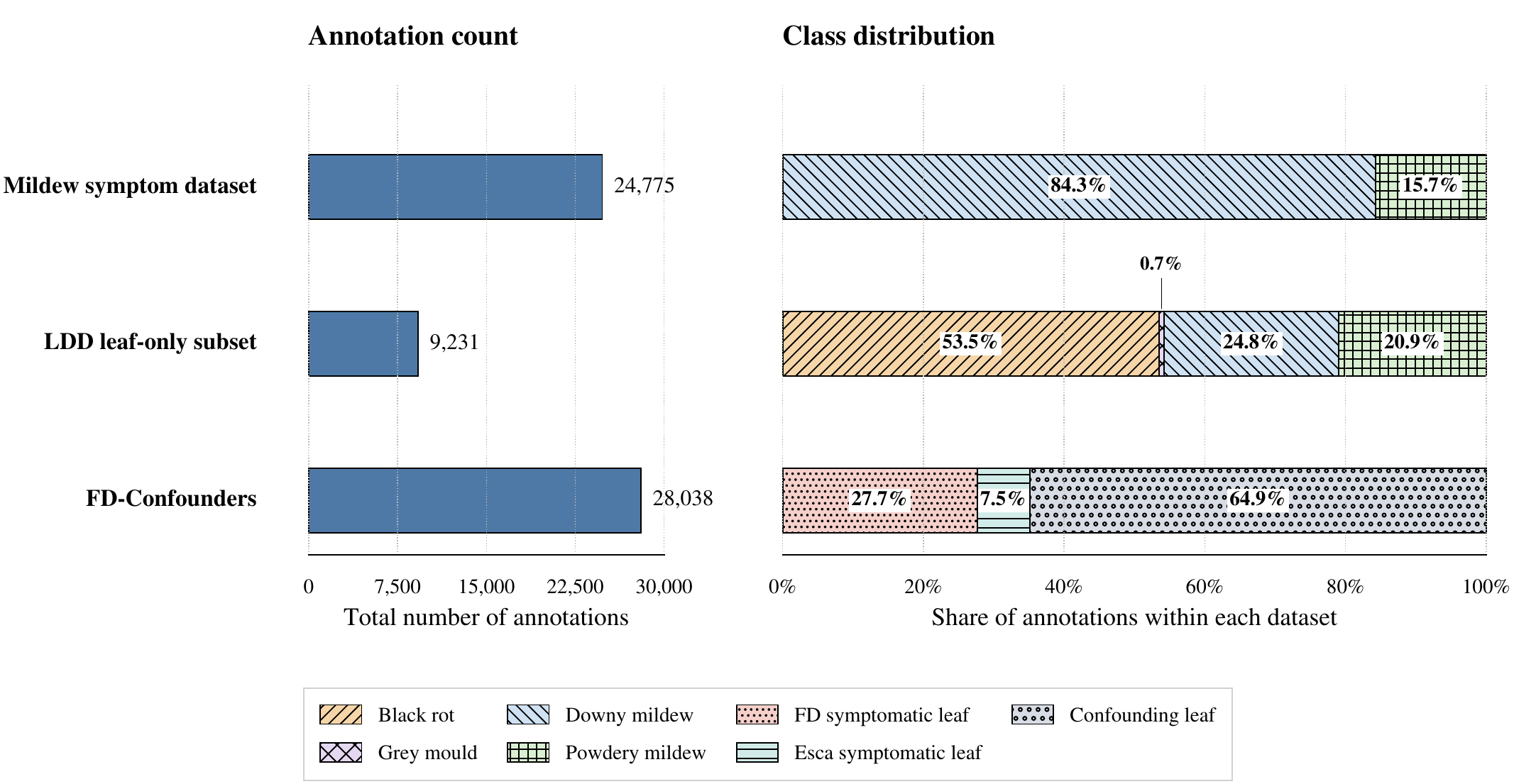}
\caption{Annotation count and class distribution of the object-detection datasets included in the benchmark. The left panel reports the total number of retained object annotations in each processed dataset, aggregated across the training, validation, and test partitions. The right panel shows the percentage distribution of annotations among the corresponding object classes. Colors and hatch patterns identify the classes and preserve readability in grayscale reproduction. The figure highlights substantial differences in dataset scale and annotation balance, including the strong dominance of downy mildew in the Mildew symptom dataset, black rot in the LDD leaf-only subset, and confounding leaves in FD-Confounders.}
\label{fig:detection_class_distributions}
\end{figure}

Figure~\ref{fig:detection_class_distributions} complements the dataset summary by showing both the total number of retained object annotations and their class distribution across the three detection datasets. The left panel illustrates substantial differences in annotation scale. FD-Confounders is the largest processed detection dataset, containing 28,038 annotated leaf instances, followed by the Mildew symptom dataset with 24,775 symptom-region annotations and the LDD leaf-only subset with 9,231 retained disease-related annotations. These values refer to object instances aggregated across the training, validation, and test partitions rather than to the number of source images.

The right panel reveals pronounced differences in class balance and annotation composition. In the Mildew symptom dataset, downy mildew accounts for 20,881 annotations, corresponding to 84.3\% of all objects, whereas powdery mildew represents only 15.7\%. The LDD leaf-only subset is also imbalanced, although its annotations span four classes. Black rot is the dominant category with 4,937 instances, or 53.5\% of the dataset, followed by downy mildew with 24.8\% and powdery mildew with 20.9\%. Grey mould is strongly underrepresented, with only 69 annotated instances, corresponding to approximately 0.7\% of all retained annotations.

FD-Confounders exhibits a different form of imbalance. Confounding leaves constitute 18,184 of the 28,038 annotations, or 64.9\% of the dataset. Flavescence dorée-symptomatic leaves represent 27.7\%, while Esca-symptomatic leaves account for only 7.5\%. This distribution reflects the purpose of the dataset, which is not limited to detecting symptomatic leaves but also requires the model to distinguish disease symptoms from visually similar non-target conditions.

These differences are important for interpreting the detection results. Datasets with a dominant class may produce aggregate mAP values that are influenced disproportionately by the most frequent category, while rare classes may remain difficult to learn and evaluate reliably. The figure therefore highlights that the three datasets differ not only in scene complexity and annotation unit, but also in the amount of supervision available for each class. Consequently, model performance should be interpreted within the annotation structure of each dataset rather than as a direct comparison of absolute task difficulty.

Overall, the selected detection datasets provide complementary perspectives on grapevine disease localization. The Mildew symptom dataset emphasizes dense local symptom detection, LDD provides multiple leaf-disease categories with instance-level spatial annotations, and FD-Confounders evaluates complete-leaf localization and discrimination in complex whole-vine scenes. This heterogeneity increases the practical relevance of the benchmark but also requires dataset-specific interpretation of detection metrics, particularly when annotation units and disease taxonomies differ.

\subsection{Dataset Characteristics and Benchmark Implications}
\label{subsec:dataset_challenges}

The preceding analysis leads to several design decisions for the benchmark. First, datasets are assigned to image-level classification, region-level classification, or object detection according to both their annotation format and the semantic meaning and completeness of those annotations. Spatial annotations that identify selected regions are not automatically treated as exhaustive object-detection ground truth.

Second, quantitative comparisons are performed primarily within each dataset. The datasets differ in acquisition environment, disease taxonomy, annotation unit, target scale, image resolution, and class balance; therefore, absolute accuracy or mAP values from different datasets should not be interpreted as direct measures of relative dataset difficulty. Region-level classification accuracy is also reported separately from complete-image classification accuracy because localization is supplied in advance.

Third, cross-dataset experiments are restricted to datasets with compatible label spaces. Even after class-name harmonization, transfer results are interpreted in relation to acquisition conditions, dataset provenance, possible image overlap, annotation semantics, and target granularity. This is particularly important for object detection, where two datasets may share disease names while defining substantially different bounding-box targets.

Finally, dataset partitions are constructed or retained with attention to source-image grouping, duplicate content, class distribution, and official evaluation protocols. Aggregate results are interpreted together with the corresponding class distributions because rare categories may receive substantially less training supervision and may be represented by relatively few evaluation instances. Based on these considerations, the following section formally defines the three evaluation settings, the dataset-harmonization procedures, the evaluated model families, and the within- and cross-dataset comparison protocols.

\section{Benchmark Methodology}
\label{sec:methodology}

This section presents the methodological design of the proposed dataset-centric benchmark. The objective of the study is not to introduce a new deep learning architecture, but to evaluate representative classification and detection models under a unified framework for grape leaf disease recognition. In line with the related-work analysis, the benchmark emphasizes lightweight and moderately sized models, since practical vineyard-monitoring systems are often expected to operate on mobile devices, embedded platforms, or edge-computing hardware with limited computational resources.

The first setting is image-level classification, where each complete input image is assigned to a disease category or to a healthy class. The second is region-level classification, where manually annotated regions are extracted from their source images and treated as classification samples. In this setting, localization is supplied by the ground-truth annotation. The third setting is object detection, where a model processes the complete image and jointly predicts the locations and categories of diagnostically relevant leaves, lesions, disease spots, or symptomatic regions.

These settings represent different input units and practical objectives. Image-level classification is suitable when one leaf or disease condition dominates the image, region-level classification evaluates discrimination after localization has already been provided, and object detection is more appropriate for field scenes containing multiple leaves, complex backgrounds, and spatially distributed symptoms. Their results are therefore reported and interpreted separately.

\subsection{Problem Formulation}
\label{subsec:problem_formulation}

This study considers three complementary evaluation settings: image-level classification, region-level classification, and object detection.

\subsubsection{Image-Level Classification}

For image-level classification, given a complete input image $x_i$, the goal is to predict its corresponding class label $y_i$:

\begin{equation}
f_{\theta}^{\mathrm{img}}(x_i) \rightarrow y_i,
\end{equation}

where $f_{\theta}^{\mathrm{img}}$ denotes an image-classification model with parameters $\theta$. The predicted label belongs to a finite dataset-specific set of categories:

\begin{equation}
y_i \in \mathcal{C}_{\mathrm{img}}
=
\{c_1,c_2,\ldots,c_K\},
\end{equation}

where $K$ is the number of image-level classes. Depending on the dataset, the class set may contain healthy leaves and disease categories such as black rot, Esca or black measles, leaf blight, downy mildew, powdery mildew, bacterial leaf spot, brown spot, or mite-related symptoms.

\subsubsection{Region-Level Classification}

For region-level classification, an annotated region $a_{ij}$ is extracted from source image $x_i$, where $j$ identifies one of the annotated regions in that image. The corresponding crop is defined as

\begin{equation}
r_{ij} = g(x_i,a_{ij}),
\end{equation}

where $g(\cdot)$ denotes the cropping and preprocessing operation. A classification model then predicts the region label $z_{ij}$:

\begin{equation}
f_{\theta}^{\mathrm{reg}}(r_{ij}) \rightarrow z_{ij},
\end{equation}

with

\begin{equation}
z_{ij} \in \mathcal{C}_{\mathrm{reg}}
=
\{c^{\mathrm{reg}}_1,c^{\mathrm{reg}}_2,\ldots,
c^{\mathrm{reg}}_{K_{\mathrm{reg}}}\},
\end{equation}

where $\mathcal{C}_{\mathrm{reg}}$ is the dataset-specific set of region-level categories and $K_{\mathrm{reg}}$ is the corresponding number of classes. In the region-level dataset considered in this benchmark, the categories correspond to healthy regions, dead arm, downy mildew, and powdery mildew. The model receives a manually localized region rather than a complete source image and is therefore evaluated only on region discrimination after localization has already been provided. To avoid information leakage, dataset partitioning is performed at the original-image level before crop extraction, ensuring that all regions derived from the same source image remain in the same training, validation, or test subset.

\subsubsection{Object Detection}

For object detection, given an input image $x_i$, the goal is to predict a set of bounding boxes, class labels, and confidence scores:

\begin{equation}
f_{\theta}^{\mathrm{det}}(x_i)
\rightarrow
\{(b_j,c_j,s_j)\}_{j=1}^{N_i},
\end{equation}

where $b_j$ is the predicted bounding box, $c_j$ is the corresponding class label, $s_j$ is the confidence score, and $N_i$ is the number of predicted instances in image $x_i$. Depending on the annotation protocol, the detection targets may correspond to complete diagnostically relevant leaves, individual lesions, disease spots, or local symptomatic regions.

Only datasets with sufficiently consistent and comprehensive spatial annotations are included in the quantitative object-detection benchmark. Segmentation polygons, where available, are converted to their enclosing bounding boxes to provide a common detection representation.

Because the three settings differ in their input unit, annotation type, and evaluation objective, their results are reported separately. Image-level classification evaluates disease recognition from complete images, region-level classification evaluates discrimination among manually localized regions, and object detection evaluates both localization and classification of spatially annotated targets.

\subsection{Classification Models}
\label{subsec:classification_models}
Representative image-classification architectures are evaluated for both image-level and region-level grape leaf disease classification. The model set was intentionally limited to lightweight and moderately sized architectures available in the \texttt{timm} library~\cite{rw2019timm}. Rather than exhaustively evaluating all available classifiers, the selected models provide representative coverage of residual CNNs, efficient CNNs, mobile-oriented CNNs, modern convolutional networks, vision transformers, hierarchical transformers, and hybrid convolutional--transformer architectures.

Table~\ref{tab:classification_models} summarizes the selected models and their approximate architectural complexity. Parameter counts and GMACs are included to characterize the scale of the architectures, but classification models are ranked quantitatively using test accuracy. The comparison is therefore intended to examine whether different architectural families provide consistent advantages across datasets with different visual and statistical characteristics.

\begin{table}[H]
\centering
\caption{Summary of the representative image classification models evaluated in this study.
The selected architectures cover residual, efficient, mobile-oriented, modern convolutional,
transformer-based, and hybrid CNN--Transformer model families. Parameter counts are approximate,
while computational complexity is reported in GMACs (giga multiply--accumulate operations) for
a single $224 \times 224$ RGB input image. The final column indicates the intended role of each
architecture within the comparative benchmark.}
\label{tab:classification_models}

\small
\setlength{\tabcolsep}{2.5pt}
\renewcommand{\arraystretch}{1.15}

\begin{adjustbox}{max width=\linewidth}
\begin{tabular}{
    >{\raggedright\arraybackslash}p{3.2cm}
    >{\raggedright\arraybackslash}p{2.7cm}
    >{\centering\arraybackslash}m{1.55cm}
    >{\centering\arraybackslash}m{1.70cm}
    >{\raggedright\arraybackslash}p{5.2cm}
}
\toprule

\textbf{Model} &
\makecell[l]{\scriptsize\textbf{Architecture}\\[-1pt]
             \scriptsize\textbf{family}} &
\makecell[c]{\scriptsize\textbf{\#Parameters}} &
\makecell[c]{\scriptsize\textbf{GMACs}\\[-1pt]
             \scriptsize\textbf{($224\times224$)}} &
\makecell[l]{\scriptsize\textbf{Role in benchmark}} \\

\midrule

ResNet-18 \cite{he2016identity} & Residual CNN & $\sim$11.7M & $\sim$1.8 & Lightweight residual CNN baseline. \\
ResNet-50 \cite{he2016identity} & Residual CNN & $\sim$25.6M & $\sim$4.1 & Standard convolutional baseline with residual learning. \\
EfficientNet-B0 \cite{tan2019efficientnet} & Efficient CNN & $\sim$5.3M & $\sim$0.4 & Compact model for accuracy--complexity analysis. \\
EfficientNet-B3 \cite{tan2019efficientnet} & Efficient CNN & $\sim$12.2M & $\sim$1.02 & Higher-capacity EfficientNet variant. \\
MobileNetV3-Large \cite{howard2019searching} & Mobile CNN & $\sim$5.5M & $\sim$0.2 & Lightweight model for mobile and edge deployment. \\
ConvNeXt-Tiny \cite{liu2022convnet} & Modern CNN & $\sim$28.6M & $\sim$4.5 & Modern convolutional baseline with transformer-inspired design. \\
ViT-S/16 \cite{dosovitskiy2020image} & Vision Transformer & $\sim$22.1M & $\sim$4.3 & Compact transformer baseline using patch-based self-attention. \\
Swin-Tiny \cite{liu2021swin} & Hierarchical ViT & $\sim$28.3M & $\sim$4.5 & Window-based transformer baseline for multi-scale visual representation. \\
DeiT-S/16 \cite{touvron2021training} & Vision Transformer & $\sim$22.1M & $\sim$4.6 & Data-efficient transformer baseline for smaller training datasets. \\
MobileViT-S \cite{mehta2021mobilevit} & Hybrid CNN--Transformer & $\sim$5.6M & $\sim$1.55 & Lightweight hybrid CNN--Transformer model for edge-oriented evaluation. \\
\bottomrule
\end{tabular}
\end{adjustbox}

\end{table}

ResNet-18 and ResNet-50 are included as residual convolutional baselines~\cite{he2016identity}. ResNet-18 provides a lightweight reference model with relatively low computational cost, while ResNet-50 provides a stronger and more widely used residual architecture. Comparing these two models allows the benchmark to analyze how increased residual-network capacity affects grape leaf disease classification performance.

EfficientNet-B0 and EfficientNet-B3 are included to evaluate compound-scaled convolutional architectures~\cite{tan2019efficientnet}. EfficientNet models balance network depth, width, and input resolution, making them suitable for tasks where both accuracy and computational efficiency are important. EfficientNet-B0 represents a compact version with low computational cost, while EfficientNet-B3 provides a higher-capacity alternative.

MobileNetV3-Large is selected as a mobile-oriented architecture~\cite{howard2019searching}. It is designed for efficient inference on resource-constrained devices through lightweight convolutional operations and optimized architectural components. Its inclusion reflects the increasing interest in deploying grape leaf disease recognition systems on smartphones, embedded boards, and other field devices.

ConvNeXt-Tiny is included as a modern convolutional architecture~\cite{liu2022convnet}. ConvNeXt incorporates several design choices inspired by transformer-based models while retaining the locality and efficiency advantages of convolutional networks. It therefore provides a useful comparison between classical CNNs, efficient CNNs, and transformer-based architectures.

ViT-S/16 is included as a compact Vision Transformer baseline~\cite{dosovitskiy2020image}. Unlike CNN-based models, Vision Transformers divide the image into patches and use self-attention to model relationships between them. This may be useful for capturing disease patterns distributed across different parts of the leaf. In the implementation, ViT-S/16 denotes the small Vision Transformer configuration using $16 \times 16$ image patches and corresponds to the \texttt{timm} model variant \texttt{vit\_small\_patch16\_224}, where 224 indicates the input resolution of $224 \times 224$ pixels.

DeiT-S/16 is included as a data-efficient transformer baseline~\cite{touvron2021training}. DeiT follows the Vision Transformer design but was developed to improve transformer training efficiency on image classification tasks. Its inclusion is relevant because grape leaf disease datasets are often relatively small, making data efficiency an important consideration. DeiT-S/16 denotes the small Data-efficient Image Transformer configuration using $16 \times 16$ image patches and is implemented with the \texttt{timm} variant \texttt{deit\_small\_patch16\_224}, where 224 indicates an input resolution of $224 \times 224$ pixels.

Swin-Tiny is included as a hierarchical transformer architecture~\cite{liu2021swin}. It uses shifted-window attention to reduce computational complexity while preserving the ability to model local and multi-scale spatial relationships. This is relevant for grape leaf disease images, where symptoms may appear at different scales and locations. In the implementation, Swin-T denotes the tiny Swin Transformer configuration and corresponds to the \texttt{timm} model variant \texttt{swin\_tiny\_patch4\_window7\_224}, which uses $4 \times 4$ input patches, self-attention within local $7 \times 7$ windows, and an input resolution of $224 \times 224$ pixels.

MobileViT-S is included as a lightweight hybrid CNN--Transformer model~\cite{mehta2021mobilevit}. It combines convolutional operations with transformer-style global representation learning, providing a useful bridge between mobile CNNs and pure transformer architectures. Its inclusion is particularly relevant for edge-oriented grape disease recognition, where both compactness and global context modeling may be beneficial. In the implementation, MobileViT-S denotes the small MobileViT configuration and corresponds to the \texttt{timm} model variant \texttt{mobilevit\_s}. This lightweight hybrid architecture combines convolutional layers for local feature extraction with transformer blocks for modeling long-range dependencies.

Within each classification setting and dataset, all architectures are evaluated using the same partitions, input pre-processing, augmentation, optimization procedure, checkpoint-selection criterion, and evaluation metric. This controls the experimental protocol within each dataset, while the use of multiple datasets allows the influence of acquisition conditions, class distributions, and dataset provenance to be examined separately from architectural differences.

\subsection{Qualitative Classification Interpretation}
\label{subsec:qualitative_interpretation}
To complement the quantitative image-level classification evaluation, Grad-CAM is used to examine the spatial evidence contributing to a selected prediction. Grad-CAM weights convolutional feature maps according to the gradients of a target class and produces a coarse class-discriminative activation map. The resulting map is overlaid on the corresponding input image to indicate which regions contributed most strongly to the selected class score.

ResNet-50 is used as the representative visualization backbone because its final convolutional stage provides a direct and well-established target for Grad-CAM. This selection is independent of the quantitative model ranking. The analysis is applied to a correctly classified GL-Portugal test image and is used only as a qualitative illustration of model behaviour. The activation map is not treated as a symptom segmentation, localization ground truth, or additional quantitative evaluation metric.

\subsection{Object Detection Models}
\label{subsec:detection_models}

For object-level grape leaf disease detection, five representative models were selected to compare detector generation, model capacity, and architectural paradigm. The benchmark focuses primarily on nano- and small-scale YOLO-family detectors, providing comparisons across YOLO generations and between two capacity levels of the latest generation. RF-DETR-Nano is included as a contrasting transformer-based detector that formulates detection through query-based set prediction rather than YOLO-style dense multiscale prediction.

The final model set comprises YOLOv8n, YOLO11n, YOLO26n, YOLO26s, and RF-DETR-Nano. YOLOv8n, YOLO11n, and YOLO26n form a controlled comparison across three YOLO generations at the nano scale. This design makes it possible to investigate whether architectural changes introduced by newer generations improve disease detection without substantially increasing model size. YOLO26s is included as a higher-capacity variant of the latest YOLO generation, allowing the effect of model scaling to be examined while retaining the same underlying architecture. RF-DETR-Nano provides a contrasting transformer-based detector and broadens the comparison beyond the convolution-oriented YOLO family. Table~\ref{tab:detection_models} summarizes the parameter counts, computational complexity, and intended role of the selected detectors.

\begin{table}[H]
\centering
\caption{Characteristics of the object detection models evaluated in the benchmark.
\textit{\#Parameters} reports the approximate number of trainable parameters, while
\textit{GFLOPs} reports the computational complexity of one forward pass for a
$1,024\times1,024$ RGB input image. Models were profiled using their fused inference
configurations. The final column describes the experimental role of each detector.}
\label{tab:detection_models}

\small
\setlength{\tabcolsep}{3pt}
\renewcommand{\arraystretch}{1.15}

\begin{adjustbox}{max width=\linewidth}
\begin{tabular}{
    >{\raggedright\arraybackslash}p{3.0cm}
    >{\centering\arraybackslash}m{1.8cm}
    >{\centering\arraybackslash}m{1.4cm}
    >{\raggedright\arraybackslash}p{8.0cm}
}
\toprule

\textbf{Model} &
{\scriptsize\textbf{\#Parameters}} &
{\scriptsize\textbf{GFLOPs}} &
\textbf{Role in benchmark} \\

\midrule

YOLOv8n & $\sim$3.2M & $\sim$22.4 & Established lightweight YOLO baseline for evaluating efficient single-stage detection. \\
YOLO11n & $\sim$2.6M & $\sim$17.0 & More recent nano-scale YOLO model used to assess generational architectural improvements under a similar computational budget. \\
YOLO26n & $\sim$2.4M & $\sim$14.5 & Latest nano-scale YOLO variant used to evaluate end-to-end detection and deployment-oriented architectural changes. \\
YOLO26s & $\sim$9.5M & $\sim$54.6 & Higher-capacity YOLO26 variant used to assess the effect of model scaling within the latest YOLO generation. \\
RF-DETR-Nano & $\sim$30.5M & $\sim$193.2 & Transformer-based detection baseline used to contrast query-based global reasoning with YOLO-style dense multiscale prediction. \\
\bottomrule
\end{tabular}
\end{adjustbox}

\end{table}

Despite belonging to the same general family, the selected YOLO generations differ in their feature-extraction blocks, attention mechanisms, bounding-box regression, and inference procedures. All three generations follow the general backbone--neck--head organization commonly used by modern single-stage detectors. The backbone extracts progressively more abstract visual features, the neck combines information across multiple spatial resolutions, and the detection head predicts object classes and bounding boxes from the resulting feature maps. Multiscale prediction is particularly relevant to the datasets considered in this study because their targets range from small localized symptoms to complete diagnostically relevant leaves.

YOLOv8n is included as the established lightweight reference model~\cite{sohan2024review}. Its backbone and neck are based primarily on C2f modules, which extend cross-stage partial feature processing by concatenating multiple intermediate feature representations. This promotes feature reuse while maintaining a relatively compact architecture. Spatial Pyramid Pooling--Fast is used to aggregate information over different effective receptive fields, while a feature-pyramid and path-aggregation structure combines low-level spatial detail with higher-level semantic information. YOLOv8 employs an anchor-free, decoupled detection head, with separate branches for bounding-box regression and class prediction. It uses Distribution Focal Loss for box-coordinate estimation and applies non-maximum suppression during inference to remove redundant predictions.

YOLO11n represents the next lightweight YOLO generation considered in the benchmark~\cite{khanam2024yolov11}. Compared with YOLOv8, it replaces the principal C2f modules with C3k2 blocks. These blocks retain cross-stage feature aggregation but allow more flexible internal bottleneck and kernel configurations. YOLO11 also introduces a C2PSA attention module after spatial-pyramid pooling, enabling the network to model longer-range relationships between spatial features. This may be relevant in vineyard images where disease evidence can be spatially distributed across a leaf or partially obscured by surrounding foliage. Its classification branch uses more computationally efficient operations than the corresponding YOLOv8 branch. Nevertheless, YOLO11 retains an anchor-free, decoupled detection head, Distribution Focal Loss for box regression, and non-maximum suppression during inference. YOLO11n therefore provides a useful comparison for determining whether improved feature extraction and attention can benefit grape disease detection without increasing the nano-scale parameter budget.

YOLO26n and YOLO26s represent the latest YOLO generation included in the benchmark~\cite{jocher2026ultralyticsyolo26unifiedrealtime}. Their backbone and neck continue to use C3k2-based feature extraction and C2PSA attention, but the detection pipeline introduces several changes intended to simplify training, inference, and deployment. In particular, YOLO26 removes Distribution Focal Loss and directly regresses bounding-box coordinates, reducing the complexity of the detection head. It also employs a dual-head design containing one-to-many and one-to-one prediction branches. The one-to-many branch provides dense supervision during training, whereas the default one-to-one branch produces end-to-end predictions without requiring non-maximum suppression. This NMS-free inference path reduces post-processing and simplifies model export and deployment.

The YOLO26 training design also incorporates progressive loss weighting and small-target-aware label assignment. These mechanisms are relevant to the present benchmark because several datasets contain numerous small, spatially distributed symptom regions. However, their practical benefit cannot be assumed from the architecture alone and is evaluated empirically on the grape disease datasets.

YOLO26n and YOLO26s share the same overall architectural design but differ in network scaling. The nano variant uses fewer channels and repeated blocks, resulting in the smallest parameter count and computational demand among the evaluated models. The small variant increases network width and depth, providing greater feature capacity at approximately four times the parameter
count and computational complexity. Their comparison therefore isolates the effect of model capacity within a single detector generation. Improved performance by YOLO26s would indicate that the task benefits from additional representational capacity, whereas comparable performance would favor YOLO26n for resource-constrained deployment.

RF-DETR-Nano is included as a transformer-based alternative to the YOLO-family detectors~\cite{robinson2025rf}. It uses a DINOv2 vision-transformer backbone and follows the detection-transformer paradigm, in which detection is formulated as set prediction rather than dense prediction over a fixed spatial grid. Transformer-based feature processing and object queries allow the detector to reason about relationships between candidate objects and wider image context. The model produces a fixed set of predictions and uses matching-based supervision to associate predictions with ground-truth targets. This design differs substantially from the multiscale dense-prediction strategy used by the YOLO models. The transformer architecture may provide advantages when targets are partially occluded or embedded in complex vineyard scenes because global contextual relationships can contribute to the prediction. However, RF-DETR-Nano contains substantially more parameters than the YOLO nano variants and therefore represents a different accuracy--efficiency operating point. The term ``Nano'' should consequently be interpreted relative to the RF-DETR family rather than as indicating a parameter count comparable to YOLO nano models.

Overall, the selected detector set supports three complementary comparisons. First, YOLOv8n, YOLO11n, and YOLO26n examine architectural evolution across YOLO generations at a broadly comparable nano scale. Second, YOLO26n and YOLO26s examine the effect of increasing capacity within the same detector generation. Third, RF-DETR-Nano provides a contrast between YOLO-style dense multiscale prediction and transformer-based query-driven set prediction. Parameter counts and theoretical computational complexity characterize the relative scale of the models, while the quantitative benchmark comparison is based on mAP@50 and mAP@50:95.

\subsection{Dataset Provenance and Image-Overlap Analysis}
\label{subsec:dataset_overlap_method}

Because several image-level classification datasets are augmented, repackaged, or potentially derived from earlier collections, exact image overlap is examined before interpreting cross-dataset results. Images are decoded into a common pixel representation and compared using exact decoded-image hashing. This procedure identifies identical visual content even when filenames, directory structures, or file encodings differ.

The overlap analysis is used to distinguish genuinely independent source-to-target evaluation from transfer involving shared image content. Detected overlaps do not alter the within-dataset partitions used for the main benchmark, but they are reported and considered when interpreting cross-dataset classification performance. Cross-dataset results involving substantial exact overlap are therefore not treated as equivalent to external validation on an independently acquired target domain.

\subsection{Within- and Cross-Dataset Evaluation}
\label{subsec:cross_dataset_strategy}

The benchmark evaluates both within-dataset performance and transfer across datasets. Within-dataset evaluation measures performance when the training, validation, and test samples originate from the same processed dataset and follow the same acquisition and annotation protocol. Cross-dataset evaluation instead examines whether representations learned from a source dataset remain effective when applied to a target dataset with different visual, statistical, or annotation characteristics.

Let $\mathcal{D}_s$ denote a source dataset and $\mathcal{D}_t$ a target dataset. Cross-dataset evaluation is performed using a harmonized class set

\begin{equation}
\mathcal{C}_{s,t}
=
\mathcal{C}_s \cap \mathcal{C}_t,
\end{equation}

where $\mathcal{C}_s$ and $\mathcal{C}_t$ are the retained class sets of the source and target datasets. A model is trained using only the training partition of $\mathcal{D}_s$ and is evaluated directly on the held-out test partition of $\mathcal{D}_t$, without target-domain fine-tuning or adaptation.

For image-level classification, cross-dataset experiments are conducted among datasets whose categories can be mapped to a compatible canonical taxonomy. The principal harmonized setting contains healthy leaves, black rot, Esca or black measles, and leaf blight or Isariopsis leaf spot. Source-to-target results are interpreted together with the dataset-provenance and exact-overlap analysis because transfer between derivative datasets may not represent fully independent external validation. Particular attention is given to transfer between controlled or visually standardized datasets and independently acquired field-oriented datasets.

For object detection, the Mildew symptom dataset and the filtered LDD subset are used for bidirectional cross-dataset evaluation because both contain downy mildew and powdery mildew. For each transfer direction, the source dataset is restricted to these two categories and the detector is retrained using the corresponding two-class label space. The resulting model is then evaluated directly on the target test partition.

The detection experiment intentionally retains the native annotation protocol of each dataset. The Mildew symptom dataset uses directly annotated bounding boxes around visible mildew symptoms, whereas LDD uses enclosing boxes derived from polygon annotations of local leaf symptoms. Consequently, the experiment evaluates transfer under both visual-domain shift and annotation-protocol shift. Shared class names are therefore treated as a necessary condition for evaluation, but not as evidence that the two datasets define identical detection targets.

Region-level classification is evaluated only within its processed dataset. Cross-dataset transfer is not performed for this setting because no second publicly available dataset in the benchmark provides a sufficiently compatible region-level label space and annotation interpretation.

By combining within-dataset evaluation, dataset-overlap analysis, and cross-dataset transfer, the methodology distinguishes strong performance within a single data source from robustness to independently acquired images and different annotation conventions.

\section{Experimental Setup}
\label{sec:experimental_setup}

The experimental protocol was designed around the three evaluation settings introduced in Section~\ref{sec:methodology}: image-level classification, region-level classification, and object detection. The image-level and region-level classification experiments used the same model-training and pre-processing pipeline, because both settings assign one class label to one input image. Their results were nevertheless evaluated separately because the image-level experiments use complete images, whereas the region-level experiments use manually localized regions extracted from annotated source images. The object-detection experiments used the original complete images and evaluated the joint localization and classification of disease targets.

Within each evaluation setting, all models were evaluated using the same dataset partitions and task-specific evaluation metrics. The classification architectures were initialized from ImageNet-pretrained weights and fine-tuned using a common two-stage protocol. The YOLO-family detectors were trained using a shared Ultralytics configuration~\cite{jocher2026ultralyticsyolo26unifiedrealtime}, whereas RF-DETR-Nano used its architecture-specific optimization procedure~\cite{robinson2025rf}. All detection models used an input resolution of $1024\times1024$ pixels. Final results were calculated only on held-out test partitions that were not used for parameter updating, hyperparameter selection, or checkpoint selection.

\subsection{Dataset Partitions}
\label{subsec:experimental_splits}
For classification datasets without predefined partitions, stratified train, validation, and test splits were constructed using proportions of 70\%, 15\%, and 15\%, respectively~\cite{kohavi1995study},~\cite{pedregosa2011scikit}. Stratification was performed according to the image-level class labels using a fixed random seed of 42. When a dataset provided official training and test partitions but no validation partition, the official test partition was retained, and a validation subset was stratified from the provided training data. Consequently, the exact global proportions of these datasets depend on the size of the original test partition. When official training, validation, and test partitions were available, they were retained without modification. The dataset-preparation pipeline additionally verified that the same source file was not assigned to more than one partition.

The same general partitioning principles were used for image-level and region-level classification. For HERMOS, the extracted regions were divided at the source-image level rather than independently at the patch level. All regions originating from the same vineyard photograph were therefore kept within the same training, validation, or test partition. This prevents visually related regions from a single source image from occurring in different partitions and producing an overly optimistic estimate of region-level generalization. The resulting source-image-grouped HERMOS partitions were then used by the classification training pipeline.

For object detection, the benchmark used the final partitions described in Section~\ref{subsubsec:object_detection_datasets}. The experiment-based partitions provided for the Mildew symptom dataset were retained. For the LDD leaf-only subset, the original validation subset was retained as the held-out test partition, while the original training subset was divided into training and validation data. FD-Confounders did not provide a predefined split for the processed leaf-detection formulation; therefore, a grouped multilabel-stratified 70/15/15 partition was constructed to preserve the distribution of object categories across the training, validation, and test subsets~\cite{sechidis2011stratification}. Duplicate and near-duplicate groups were retained within the same partition during this process.

\subsection{Classification Pre-processing and Training Protocol}
\label{subsec:classification_training}
The same pre-processing and training protocol was applied to all image-level and region-level classification models. Each model received an RGB input of $224\times224$ pixels. During training, images were first resized so that their shorter spatial dimension was 256 pixels and were then randomly cropped to $224\times224$ pixels. Data augmentation consisted of random horizontal and vertical flipping, each with a probability of 0.5; random rotation within $\pm15^{\circ}$; color jitter with brightness, contrast, and saturation factors of 0.3 and a hue factor of 0.1; and random grayscale conversion with a probability of 0.05.

Validation and test images were resized to 256 pixels and center-cropped to $224\times224$ pixels without stochastic augmentation. All images were converted to tensors and normalized using the ImageNet channel statistics:

\begin{equation}
\boldsymbol{\mu} = (0.485, 0.456, 0.406), \qquad
\boldsymbol{\sigma} = (0.229, 0.224, 0.225).
\end{equation}

The same deterministic evaluation transformation was used for within-dataset testing and cross-dataset classification. All classification architectures were initialized using ImageNet-pretrained weights, and their final classification layers were replaced according to the number of classes in the corresponding dataset. Models were trained with a batch size of 64 for a maximum of 100 epochs using a two-stage transfer-learning procedure.

During the first stage, the pretrained backbone was frozen and only the newly initialized classification head was optimized for five epochs. AdamW was used with a learning rate of $10^{-3}$ and a weight decay of $10^{-4}$. A cosine-annealing learning-rate schedule was applied, with the minimum learning rate set to $10^{-5}$.

During the second stage, all model parameters were unfrozen and jointly fine-tuned. Different learning rates were assigned to the pretrained backbone and the classification head. The backbone was optimized with a learning rate of $10^{-4}$, while the classification head retained a learning rate of $10^{-3}$. AdamW with a weight decay of $10^{-4}$ was again used, together with cosine annealing to a minimum learning rate of $10^{-6}$.

The training objective was weighted cross-entropy loss with label smoothing of 0.1:

\begin{equation}
\mathcal{L}_{\mathrm{cls}}
=
-\sum_{k=1}^{K}
w_k \widetilde{y}_{k}\log p_{k},
\end{equation}

where $K$ is the number of classes, $p_k$ is the predicted probability of class $k$, $\widetilde{y}_{k}$ denotes the label-smoothed target, and $w_k$ is the weight assigned to class $k$. Balanced class weights were calculated separately for each dataset using only the corresponding training partition. The weighting procedure was applied in every image-level and region-level classification experiment to reduce the influence of class imbalance without using information from the validation or test partitions.

Model selection was based on validation loss. The checkpoint with the lowest validation loss was retained, and early stopping was applied during full fine-tuning when the validation loss failed to improve for 20 consecutive epochs. A fixed seed of 42 was set before training each model. Four data-loading workers were used; training batches were shuffled, whereas validation and test samples were processed in deterministic order.

The same protocol was applied to region-level classification. In this setting, the model inputs were annotated regions extracted from their source images rather than complete images. No localization operation was performed during training or testing because the region coordinates were provided by the ground-truth annotations.

\subsection{YOLO Training Protocol}
\label{subsec:yolo_training}

YOLOv8n, YOLO11n, YOLO26n, and YOLO26s were trained using their corresponding pretrained Ultralytics checkpoints. Each detector was trained independently on each object-detection dataset for 100 epochs using an input resolution of $1,024\times1,024$ pixels and a batch size of 8. The random seed was fixed to 1, four data-loading workers were used, and training was performed on a single GPU. All optimizer, loss, learning-rate scheduling, and augmentation parameters not explicitly specified were retained from the corresponding Ultralytics default configuration.

For every YOLO detector, checkpoint selection was performed using the validation partition during training. After training, the checkpoint identified by the Ultralytics framework as the best model was reloaded and evaluated once on the held-out test partition. Test evaluation used a batch size of 1 and the same $1,024\times1,024$ input resolution as training. For YOLO26, evaluation used the conventional NMS-based output path by setting \texttt{end2end=False}, thereby maintaining a consistent output-processing procedure across the YOLO-family detectors. The reported mAP values were extracted from the bounding-box evaluation results of the selected checkpoint.

\subsection{RF-DETR-Nano Training Protocol}
\label{subsec:rfdetr_training}

RF-DETR-Nano was initialized from its pretrained checkpoint and adapted to the number of categories in each detection dataset. The original YOLO-format datasets were exposed to the RF-DETR pipeline through directory adapters, without modifying or duplicating the original images and annotations.

RF-DETR-Nano was trained for a maximum of 100 epochs using an input resolution of $1024\times1024$ pixels. A physical batch size of 2 and eight gradient accumulation steps were used, corresponding to an effective batch size of 16. The initial learning rate was $10^{-4}$, and four data-loading workers were used. Exponential moving average weights were enabled during training. Validation was performed after every epoch, while full checkpoints were saved at intervals of 10 epochs.

Early stopping was applied with a patience of 20 epochs and a minimum required validation improvement of 0.001. The first three epochs were excluded from best-checkpoint selection to avoid selecting a checkpoint before the pretrained detector had adapted to the target categories. The final model was obtained from the best validation checkpoint selected during training, considering both the regular and exponential-moving-average model states.

Final RF-DETR-Nano predictions were evaluated using a COCO-style bounding-box evaluation pipeline. All input images were explicitly converted to three-channel RGB format before inference. The YOLO-format reference annotations and RF-DETR predictions were converted to COCO bounding-box representations, and average precision was calculated over the required intersection-over-union thresholds. The held-out test partition was evaluated independently of the validation partition.

\subsection{Cross-Dataset Experimental Protocol}
\label{subsec:cross_dataset_experimental_setup}
Cross-dataset classification experiments used the same deterministic $224\times224$ evaluation transformation as within-dataset testing. For each source-to-target direction, a model was trained using only the training and validation partitions of the source dataset and was then evaluated directly on the held-out test partition of the target dataset. No target-domain samples were used for parameter updating, checkpoint selection, calibration, or fine-tuning. Disease names were mapped to the harmonized class taxonomy defined in Section~\ref{subsec:cross_dataset_strategy}. Swin-Tiny was used as the representative architecture for all cross-dataset classification experiments.

Exact decoded-image overlap among the classification datasets was analyzed separately and considered when interpreting the transfer results. Therefore, cross-dataset results involving shared visual content were distinguished from evaluation on independently acquired target images.

For cross-dataset object detection, the Mildew symptom dataset and the filtered LDD subset were restricted to their two shared categories: downy mildew and powdery mildew. In each transfer direction, the detector was retrained as a two-class model on the source dataset and evaluated directly on the held-out test partition of the target dataset. The target-domain annotations were not used for training, validation, confidence calibration, or model selection. The native bounding-box definition of each dataset was retained, so the experiment reflects both visual-domain shift and annotation-protocol shift. YOLO26s was used as the representative detector for the bidirectional cross-dataset object-detection experiment.

\subsection{Qualitative Interpretability Setup}
\label{subsec:gradcam_setup}

To complement the quantitative image-level classification evaluation, Grad-CAM was applied to a representative correctly classified test image from GL-Portugal. ResNet-50 was used as the visualization backbone because its final convolutional stage provides a direct and established target for class-activation mapping. The activation map was calculated for the predicted class using the final residual convolutional stage and was upsampled and overlaid on the corresponding $224\times224$ evaluation image.

The selection of ResNet-50 for this analysis was independent of the quantitative model ranking. Grad-CAM was used only to provide a qualitative illustration of spatial model attention and did not contribute to checkpoint selection, architecture ranking, or quantitative evaluation. The resulting map was not interpreted as a symptom-segmentation output.

\subsection{Evaluation Metrics}
\label{subsec:evaluation_metrics}

Image-level and region-level classification performance was evaluated using
accuracy:
\begin{equation}
\mathrm{Accuracy}
=
\frac{N_{\mathrm{correct}}}{N_{\mathrm{total}}},
\end{equation}
where $N_{\mathrm{correct}}$ is the number of correctly classified samples and $N_{\mathrm{total}}$ is the total number of samples in the held-out test partition. Although the same metric was used for both classification settings, their values are reported separately because complete images and manually localized regions represent different input units and evaluation objectives.

Object-detection performance was evaluated using mAP@50 and mAP@50:95. The former calculates mean average precision at an intersection-over-union threshold of 0.5. The latter averages class-specific average precision over intersection-over-union thresholds from 0.5 to 0.95 in increments of 0.05, providing a stricter assessment of both classification and bounding-box localization. Metrics were calculated independently for each dataset because the datasets differ in disease taxonomy, target scale, scene complexity, class distribution, and annotation unit.

\subsection{Implementation Details}
\label{subsec:implementation_details}

The classification experiments were implemented in Python using PyTorch, Torchvision, and the \texttt{timm} model library. Dataset partitioning, class-weight calculation, and classification metric computation used scikit-learn. YOLO training and evaluation were implemented using the Ultralytics framework, while RF-DETR-Nano was trained using RF-DETR version~1.8.3. COCO-style evaluation of RF-DETR predictions was performed using \texttt{pycocotools}. The experiments were conducted on a workstation equipped with an NVIDIA GeForce RTX 5060 Ti GPU with 16~GB of memory.

The classification experiments used a batch size of 64, the YOLO models used a batch size of 8, and RF-DETR-Nano used a physical batch size of 2 with gradient accumulation to obtain an effective batch size of 16. Classification and YOLO experiments used the fixed random seeds reported above. Each model--dataset combination, including each cross-dataset transfer direction, was trained once. Model checkpoints, training histories, dataset-split manifests, overlap-analysis outputs, and final metric files were retained to support reproducibility.

\section{Results and Discussion}
\label{sec:results}

\subsection{Image-level Classification Results}
\label{subsec:classification_results}
This section presents the image-level classification results using accuracy as the evaluation metric. The analysis is divided into four parts. First, we report within-dataset accuracy, where training, validation, and testing are performed on the same dataset. Second, we examine class-level accuracy for the two lowest-performing datasets. Third, we analyze exact image overlap among datasets to support the interpretation of the results. Fourth, we evaluate cross-dataset accuracy for datasets sharing a compatible four-class label space.

\subsubsection{Within-Dataset Accuracy}
\label{subsubsec:within_dataset_accuracy}

Table~\ref{tab:classification_accuracy_matrix} reports the within-dataset classification accuracy of the evaluated models across all grape leaf disease classification datasets. The results show that several datasets are almost saturated under standard within-dataset evaluation. PlantVillage, PDR2018, the Grape Leaf Disease Dataset, the Grape Leaf Disease Augmented Dataset, and the New Plant Diseases grape subset achieve near-perfect or perfect accuracy for most evaluated architectures. In these cases, the choice of model architecture has only a limited effect on the final result, because even compact models are able to learn the visual patterns present in the datasets.

\begin{table}[t]
\centering
\caption{Within-dataset classification accuracy across grape leaf disease datasets.
Rows correspond to datasets and columns correspond to classification models.}
\label{tab:classification_accuracy_matrix}

\scriptsize
\setlength{\tabcolsep}{3.0pt}
\renewcommand{\arraystretch}{1.12}

\begin{adjustbox}{max width=\linewidth}
\begin{tabular}{
    >{\raggedright\arraybackslash}p{2.8cm}
    *{10}{c}
}
\toprule

\textbf{Dataset} &
\rotatebox{60}{\textbf{ViT-S/16}} &
\rotatebox{60}{\textbf{ResNet-50}} &
\rotatebox{60}{\textbf{ResNet-18}} &
\rotatebox{60}{\textbf{EffNet-B0}} &
\rotatebox{60}{\textbf{EffNet-B3}} &
\rotatebox{60}{\textbf{MobileNetV3}} &
\rotatebox{60}{\textbf{ConvNeXt-T}} &
\rotatebox{60}{\textbf{Swin-T}} &
\rotatebox{60}{\textbf{MobileViT-S}} &
\rotatebox{60}{\textbf{DeiT-S/16}} \\

\midrule

GVLiD
& 0.8946 & 0.8697 & 0.8659 & 0.8736 & 0.8678
& 0.8831 & 0.8812 & 0.8966 & 0.8889 & 0.8966 \\

NGLD
& 0.9976 & 0.9976 & 0.9976 & 0.9878 & 0.9951
& 0.9976 & 0.9976 & 0.9976 & 0.9976 & 0.9976 \\

PlantVillage
& 1.0000 & 1.0000 & 0.9967 & 1.0000 & 0.9984
& 1.0000 & 1.0000 & 0.9984 & 1.0000 & 0.9984 \\

PDR2018
& 0.9950 & 0.9950 & 0.9975 & 0.9975 & 0.9975
& 1.0000 & 0.9975 & 0.9975 & 0.9950 & 0.9975 \\

GLDD
& 1.0000 & 1.0000 & 1.0000 & 0.9875 & 1.0000
& 0.9938 & 0.9938 & 1.0000 & 1.0000 & 1.0000 \\

GLDD Augmented
& 0.9997 & 0.9983 & 0.9983 & 0.9994 & 0.9989
& 0.9992 & 1.0000 & 0.9994 & 0.9994 & 0.9983 \\

GLHCD
& 0.8456 & 0.8404 & 0.8016 & 0.8530 & 0.8530
& 0.8452 & 0.8504 & 0.8523 & 0.8452 & 0.8445 \\

New Plant Diseases
& 1.0000 & 1.0000 & 1.0000 & 1.0000 & 1.0000
& 1.0000 & 1.0000 & 1.0000 & 1.0000 & 1.0000 \\

GL-Portugal
& 0.9684 & 0.9583 & 0.9583 & 0.9709 & 0.9633
& 0.9646 & 0.9823 & 0.9836 & 0.9722 & 0.9697 \\

\bottomrule
\end{tabular}
\end{adjustbox}

\end{table}

The New Plant Diseases grape subset is the most saturated dataset, with all evaluated models achieving 1.0000 accuracy. Similar behavior is observed for PlantVillage and the Grape Leaf Disease Dataset, where multiple architectures also reach perfect accuracy. PDR2018 and the Grape Leaf Disease Augmented Dataset likewise produce near-perfect results across the evaluated model families. These findings indicate that controlled, visually standardized, augmented, or derivative grape leaf disease datasets may provide limited discrimination between modern pretrained architectures.

GL-Portugal represents an important intermediate case between controlled imagery and fully unconstrained field conditions. Although its images were acquired in vineyard environments under natural illumination, the dataset authors report that the photographed leaves were fully visible in most cases. This acquisition procedure improves symptom visibility but reduces scene complexity compared with real-time vineyard monitoring, where symptomatic leaves are frequently partially occluded within the canopy. Accordingly, all evaluated models achieve high accuracy on GL-Portugal, ranging from 0.9583 for ResNet-50 and ResNet-18 to 0.9836 for Swin-Tiny, while ConvNeXt-Tiny reaches 0.9823. The relatively narrow difference between the lowest- and highest-performing models indicates that the dataset is consistently handled well across architecture families. However, these results should not be interpreted as evidence of equivalent robustness under fully unconstrained vineyard conditions. The predominantly leaf-centered framing, limited occlusion, balanced class distribution, and clearly visible symptom patterns may partly explain the high within-dataset accuracy.

In contrast, GVLiD and GLHCD are substantially more challenging. For GVLiD, the highest accuracy is 0.8966, achieved by Swin-Tiny and DeiT-S/16, while GLHCD reaches a maximum accuracy of 0.8530 with EfficientNet-B0 and EfficientNet-B3. The contrast between GVLiD and GL-Portugal is particularly informative. Both datasets were acquired in vineyard environments, but they differ in image composition and scene complexity. GL-Portugal predominantly presents fully visible individual leaves with clearly observable symptoms, whereas GVLiD contains richer field variability and more complex image conditions. This comparison indicates that the broad designation ``field-acquired'' is insufficient to characterize dataset difficulty; leaf visibility, occlusion, framing, background complexity, and symptom prominence must also be considered.

Overall, the within-dataset results show that accuracy is strongly influenced by the specific acquisition and curation protocol. Near-perfect performance is not restricted to laboratory or controlled-background datasets, as demonstrated by GL-Portugal. Nevertheless, its leaf-centered acquisition procedure reduces some of the complexity encountered in operational vineyard monitoring. Dataset difficulty should therefore be interpreted according to the combined effects of acquisition environment, leaf visibility, occlusion, image framing, background complexity, disease taxonomy, symptom separability, and dataset construction.

\subsubsection{Class-Level Accuracy on Challenging Datasets}
\label{subsubsec:class_level_accuracy}
Although the main comparison is based on overall accuracy, dataset-level accuracy can hide differences between disease categories. Therefore, we additionally report class-level accuracy for the two lowest-performing datasets, GVLiD and GLHCD. For each class, class-level accuracy is computed as the proportion of correctly classified test images from that class. GL-Portugal was not included in this focused analysis because all evaluated architectures achieved between 0.9583 and 0.9836 overall accuracy, whereas the purpose of this subsection is to identify the classes responsible for the substantially lower performance observed on the most challenging datasets.

For GVLiD, Swin-Tiny was selected because it achieved the highest overall accuracy together with DeiT-S/16. For GLHCD, EfficientNet-B0 was selected because it achieved the highest overall accuracy while also being one of the most compact models. Table~\ref{tab:classification_class_level_accuracy} summarizes the class-level results.

\begin{table}[t]
\caption{Class-level accuracy on the two most challenging classification datasets.}
\label{tab:classification_class_level_accuracy}
\centering
\small
\setlength{\tabcolsep}{6pt}
\renewcommand{\arraystretch}{1.12}

\begin{tabular}{llcc}
\toprule
\textbf{Dataset} & \textbf{Class} & \textbf{Selected model} & \textbf{Class-level accuracy} \\
\midrule
GVLiD & Healthy & Swin-Tiny & 1.0000 \\
GVLiD & Black rot & Swin-Tiny & 1.0000 \\
GVLiD & Esca / Black measles & Swin-Tiny & 0.7388 \\
GVLiD & Leaf blight & Swin-Tiny & 0.8119 \\
\midrule
GLHCD & Healthy & EfficientNet-B0 & 0.9097 \\
GLHCD & Black rot & EfficientNet-B0 & 0.8870 \\
GLHCD & Downy mildew & EfficientNet-B0 & 0.8810 \\
GLHCD & Brown spot & EfficientNet-B0 & 0.7827 \\
GLHCD & Mites & EfficientNet-B0 & 0.7944 \\
\bottomrule
\end{tabular}
\end{table}

The results show that the reduced accuracy on GVLiD is mainly associated with Esca/black measles and leaf blight. In contrast, healthy leaves and black rot are classified correctly in all test cases by the selected Swin-T model. This indicates that GVLiD is not uniformly difficult across all classes; rather, the main challenge comes from visually ambiguous disease categories.

For GLHCD, the class-level accuracy is more evenly distributed but remains lower than on the controlled four-class datasets. Healthy leaves achieve the highest class-level accuracy, while brown spot and mites are the most difficult classes. This suggests that the lower overall accuracy on GLHCD is caused by broader visual heterogeneity and greater similarity between some disease symptoms. These class-level results support the dataset-centric interpretation that classification performance depends not only on model architecture, but also on the visual separability of the disease categories within each dataset.

\subsubsection{Dataset Overlap Analysis}
\label{subsubsec:dataset_overlap_analysis}
Because several controlled datasets produced near-perfect within-dataset accuracy, an additional overlap analysis was performed. Exact duplicate images were identified by hashing a common decoded-pixel representation, allowing identical visual content to be detected independently of filenames and directory structure. Table~\ref{tab:classification_exact_overlap} summarizes the detected exact-image overlap among the classification datasets.

\begin{table}[t]
\caption{Exact image overlap detected among classification datasets using hashing of a common decoded-pixel representation.}
\label{tab:classification_exact_overlap}
\centering
\small
\setlength{\tabcolsep}{6pt}
\renewcommand{\arraystretch}{1.12}
\begin{tabular}{llc}
\toprule
\textbf{Dataset A} & \textbf{Dataset B} & \textbf{\#Shared exact images} \\
\midrule
PlantVillage & New Plant Diseases & 4,062 \\
GLDD & PlantVillage & 1,598 \\
GLDD & New Plant Diseases & 1,598 \\
PDR2018 & PlantVillage & 89 \\
PDR2018 & New Plant Diseases & 89 \\
GLDD & PDR2018 & 31 \\
\bottomrule
\end{tabular}
\end{table}

The overlap analysis shows that some of the controlled four-class datasets are not fully independent. In particular, the PlantVillage grape subset is completely represented within the New Plant Diseases grape subset, and the Grape Leaf Disease Dataset also overlaps completely with this controlled-dataset group. Smaller overlaps are also present between PDR2018 and the same group. This finding is important because it explains part of the near-perfect transfer observed among these datasets. Therefore, high accuracy among these controlled datasets should not be interpreted as independent evidence of generalization, but rather as evidence of shared image provenance or derivative dataset construction. No exact duplicate images were detected between GL-Portugal and the other classification datasets, supporting its treatment as an independently acquired collection.

\subsubsection{Cross-Dataset Accuracy}
\label{subsubsec:cross_dataset_accuracy}
Cross-dataset classification was evaluated using Swin-Tiny as the representative architecture. The same model architecture and training protocol were used for every source--target combination so that differences in transfer performance could be attributed primarily to dataset shift rather than to changes in model architecture. Swin-Tiny was selected because it showed consistently strong within-dataset performance across the classification benchmark while representing a modern hierarchical transformer architecture.

Table~\ref{tab:classification_cross_dataset_accuracy} reports the resulting cross-dataset accuracy. Rows indicate the source dataset used for training, while columns indicate the target dataset used for testing. The experiment is limited to datasets with the compatible four-class label space consisting of healthy leaves, black rot, Esca or black measles, and leaf blight or Isariopsis leaf spot. NGLD, GLHCD, and GL-Portugal are excluded because their complete label spaces are not directly compatible with this four-class configuration. Although GL-Portugal partially overlaps with the common taxonomy, its remaining categories prevent a direct four-class source--target evaluation without discarding a substantial portion of the dataset.

\begin{table}[H]
\centering
\caption{Cross-dataset classification accuracy obtained using Swin-Tiny. Rows indicate the source dataset used for training and columns indicate the target dataset used for testing. The same architecture and training protocol were used for all source--target combinations. Diagonal entries correspond to within-dataset evaluation of Swin-Tiny under the cross-dataset experimental protocol.}
\label{tab:classification_cross_dataset_accuracy}

\small
\setlength{\tabcolsep}{2.3pt}
\renewcommand{\arraystretch}{1.15}

\begin{adjustbox}{max width=\linewidth}
\begin{tabular}{
    @{}
    >{\raggedright\arraybackslash}p{2.9cm}
    >{\centering\arraybackslash}m{1.35cm}
    >{\centering\arraybackslash}m{1.55cm}
    >{\centering\arraybackslash}m{1.35cm}
    >{\centering\arraybackslash}m{1.30cm}
    >{\centering\arraybackslash}m{2.20cm}
    >{\centering\arraybackslash}m{1.85cm}
    @{}
}
\toprule

\makecell[l]{\textbf{Source /}\\\textbf{Target}} &
\textbf{GVLiD} &
\makecell[c]{\textbf{Plant}\\\textbf{Village}} &
\makecell[c]{\textbf{PDR}\\\textbf{2018}} &
\textbf{GLDD} &
\makecell[c]{\textbf{GLDD}\\\textbf{Augmented}} &
\makecell[c]{\textbf{New Plant}\\\textbf{Diseases}} \\

\midrule

GVLiD
& 0.8966
& 0.2902
& 0.3017
& 0.3250
& 0.2778
& 0.2731
\\

PlantVillage
& 0.1667
& 0.9984
& 0.9975
& 1.0000
& 0.6532
& 1.0000
\\

PDR2018
& 0.1935
& 0.9902
& 0.9975
& 1.0000
& 0.5413
& 0.9945
\\

GLDD
& 0.2222
& 0.9902
& 0.9900
& 1.0000
& 0.4224
& 0.9922
\\

GLDD Augmented
& 0.3218
& 0.6098
& 0.5810
& 0.6812
& 0.9994
& 0.6842
\\

New Plant Diseases
& 0.2318
& 1.0000
& 0.9975
& 1.0000
& 0.6440
& 1.0000
\\

\bottomrule
\end{tabular}
\end{adjustbox}

\end{table}

The diagonal values in Table~\ref{tab:classification_cross_dataset_accuracy} correspond specifically to Swin-Tiny and should therefore not be interpreted as the best within-dataset result reported in Table~\ref{tab:classification_accuracy_matrix}, which compares all evaluated architectures. The cross-dataset results substantially change the interpretation of the classification benchmark. Among PlantVillage, PDR2018, the Grape Leaf Disease Dataset, and the New Plant Diseases grape subset, the transfer accuracy is very high, frequently close to or equal to 1.0000. However, based on the overlap analysis in Table~\ref{tab:classification_exact_overlap}, these results should be interpreted cautiously. The high transfer accuracy among these datasets is likely influenced by shared or derivative image content, rather than reflecting independent cross-dataset robustness.

GVLiD shows the strongest domain shift. Models trained on controlled or derivative datasets perform poorly when evaluated on GVLiD, with accuracies between 0.1667 and 0.3218. The reverse direction is also weak: the model trained on GVLiD achieves only 0.2731--0.3250 accuracy when evaluated on the controlled or augmented target datasets. This indicates that GVLiD represents a substantially different visual domain, likely due to field acquisition, natural background, variable illumination, leaf pose, symptom appearance, and image quality.

The Grape Leaf Disease Augmented Dataset shows intermediate behavior. Its diagonal accuracy is very high, but its transfer to the controlled datasets is clearly lower than the transfer observed within the overlapping controlled-dataset group. When used as a source dataset, it reaches 0.6098 on PlantVillage, 0.5810 on PDR2018, 0.6812 on the Grape Leaf Disease Dataset, and 0.6842 on the New Plant Diseases grape subset. When used as a target dataset, models trained on controlled datasets also show reduced accuracy, ranging from 0.4224 to 0.6532. This suggests that augmentation changes the visual distribution, but does not fully reproduce the variability of field-acquired vineyard images.

Taken together, the cross-dataset results show that high within-dataset accuracy is not sufficient evidence of generalization. Controlled and derivative datasets can produce excellent accuracy, but this performance does not necessarily transfer to GVLiD, which represents an independent field-acquired domain with a compatible four-class taxonomy. The large accuracy drop involving GVLiD provides clear evidence of dataset shift between this dataset and the controlled or derivative group. However, this conclusion should not be generalized to all vineyard-acquired datasets, because GL-Portugal uses a different five-class taxonomy and could not be included in the same cross-dataset matrix. Moreover, its predominantly fully visible and leaf-centered images represent less complex conditions than unconstrained canopy-level monitoring. Future experiments could examine transfer between datasets sharing downy mildew, powdery mildew, Esca, or mite-related categories after defining a sufficiently justified common label subset.

\subsubsection{Summary of Classification Findings}
\label{subsubsec:classification_summary}
The classification experiments lead to four main findings. First, several controlled, augmented, or derivative datasets produce near-perfect within-dataset accuracy, indicating that conventional train--test evaluation may provide limited discrimination between modern architectures. Second, GL-Portugal also achieves consistently high accuracy despite being acquired in vineyard environments; however, its predominantly fully visible and leaf-centered images reduce occlusion and scene complexity compared with unconstrained canopy-level monitoring. Third, exact image overlap exists among several controlled four-class datasets, which affects the interpretation of their within-group cross-dataset transfer. Fourth, the cross-dataset evaluation reveals a substantial domain gap specifically between GVLiD and the controlled or derivative dataset group. These findings show that classification performance depends on the combined effects of acquisition environment, leaf visibility, occlusion, image framing, background complexity, dataset provenance, disease taxonomy, symptom separability, and class distribution. Consequently, accuracy should be reported together with transparent descriptions of the image-acquisition protocol, overlap analysis, and cross-dataset evaluation whenever compatible label spaces are available.

\subsubsection{Qualitative Interpretation Using Grad-CAM}
\label{subsubsec:gradcam_analysis}
To complement the quantitative image-level classification results, Gradient-weighted Class Activation Mapping (Grad-CAM) was used to examine the image regions that contributed most strongly to a representative model prediction~\cite{selvaraju2017grad}. Grad-CAM constructs a class-specific activation map by weighting the spatial feature maps of a convolutional layer according to the gradients associated with the selected class. The resulting visualization does not provide a pixel-level disease annotation, but it offers a qualitative indication of the spatial evidence used by the classifier.

ResNet-50 was selected as the representative model for this qualitative analysis. This choice was not based on it achieving the highest classification accuracy in the benchmark. The quantitative comparison and model ranking remain based exclusively on the test-set results reported in the preceding sections. Rather, ResNet-50 was selected because its final convolutional feature maps provide a direct and well-established basis for Grad-CAM visualization and produced spatially coherent activation maps that were straightforward to interpret. Transformer-based models were also evaluated quantitatively, but their activation visualizations depend more strongly on the selected transformer block, token reshaping procedure, and spatial aggregation strategy. Using ResNet-50 therefore provides a clear illustrative example without implying that it is the overall best-performing architecture.

\begin{figure}[H]
\centering
\includegraphics[width=\textwidth]
{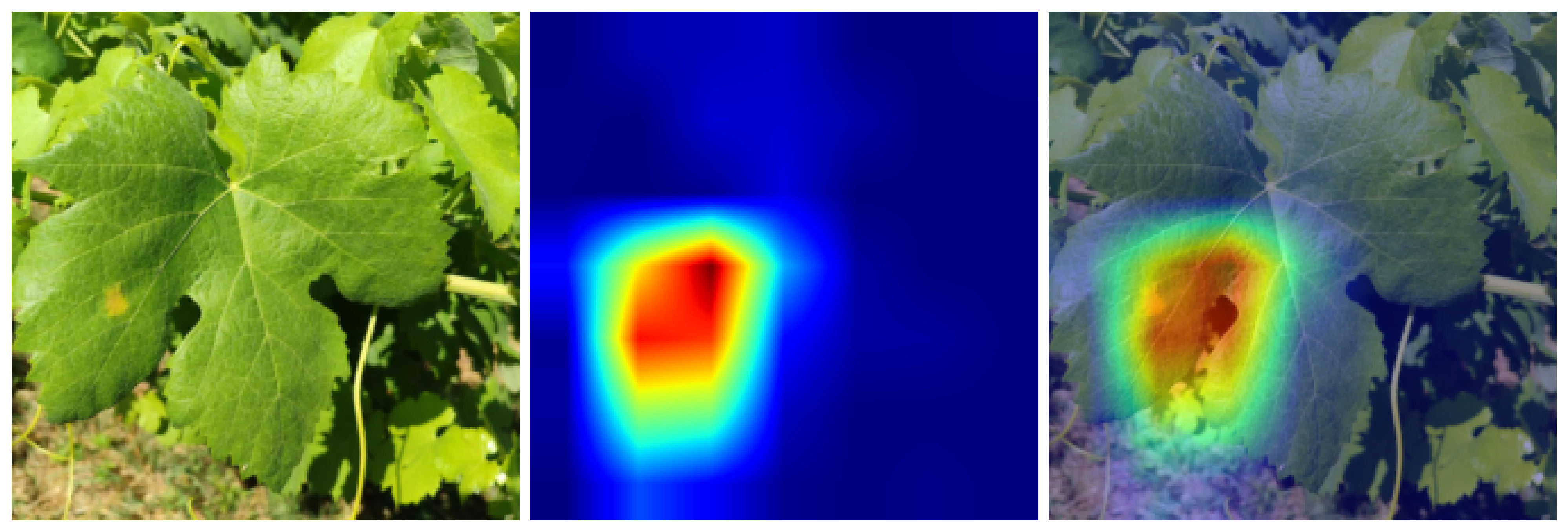}
\caption{Representative Grad-CAM visualization for a correctly classified GL-Portugal downy-mildew image using ResNet-50. From left to right: the preprocessed input image, the Grad-CAM activation map calculated for the downy-mildew class, and the activation map overlaid on the input image. The model predicted downy mildew with a probability of 92.9\%. Warmer colors indicate image regions with a stronger contribution to the selected class score. The activation is concentrated mainly on the foreground leaf and its visibly discolored region; however, the map represents coarse class-discriminative evidence and should not be interpreted as a precise symptom segmentation.}
\label{fig:gradcam_resnet50}
\end{figure}

Figure~\ref{fig:gradcam_resnet50} presents a representative correctly classified downy-mildew image. The model assigned a probability of 92.9\% to the downy-mildew class. The Grad-CAM response is concentrated primarily on the central portion of the foreground leaf, including the visibly discolored symptomatic area, rather than being distributed uniformly across the surrounding foliage and background. This suggests that the prediction was influenced by local visual evidence associated with the affected leaf rather than exclusively by broad scene context.

The activation region is nevertheless spatially coarser than the visible symptom boundary. This is expected because Grad-CAM is generated from low-resolution feature maps in the final convolutional stage and is subsequently upsampled to the input-image dimensions. Consequently, the visualization should be interpreted as an approximate class-discriminative region and not as a lesion-segmentation result. Furthermore, a single example cannot establish that the classifier consistently relies on biologically meaningful evidence across the complete test set. The figure is therefore included as a qualitative illustration of model behaviour rather than as an additional quantitative evaluation criterion.

\subsection{Region-Level Classification Results}
\label{subsec:region_level_results}

Table~\ref{tab:hermos_region_results} presents the region-level classification accuracy obtained on the HERMOS dataset. In this evaluation setting, each model receives an annotated region cropped from its original source image. The localization of the region is therefore supplied by the ground-truth annotation, and the reported accuracy measures only the model's ability to
distinguish among healthy regions, dead arm, downy mildew, and powdery mildew.

\begin{table}[H]
\caption{Region-level classification accuracy on the HERMOS held-out test partition. All regions originating from the same source image were retained in the same dataset partition to prevent information leakage. The best result is shown in bold.}
\label{tab:hermos_region_results}
\centering
\small
\begin{tabular}{l c}
\toprule
\textbf{Model} & \textbf{Accuracy} \\
\midrule
ResNet-18          & 0.8842 \\
ResNet-50          & 0.8975 \\
EfficientNet-B0    & 0.9015 \\
EfficientNet-B3    & 0.8936 \\
MobileNetV3-Large  & 0.9010 \\
ConvNeXt-Tiny      & 0.8911 \\
ViT-S/16           & 0.9054 \\
DeiT-S/16          & \textbf{0.9148} \\
Swin-Tiny          & 0.9138 \\
MobileViT-S        & 0.8837 \\
\bottomrule
\end{tabular}
\end{table}

The results are relatively concentrated, with accuracy ranging from 0.8837 for MobileViT-S to 0.9148 for DeiT-S/16, corresponding to a difference of only 3.11 percentage points across the complete model set. This relatively narrow range indicates that all evaluated architectures learn useful discriminative representations from the extracted HERMOS regions, although none of the models completely resolves the four-class recognition problem.

The strongest results are obtained by the transformer-based models. DeiT-S/16 achieves the highest accuracy of 0.9148, while Swin-Tiny follows closely with 0.9138. ViT-S/16 ranks third with an accuracy of 0.9054. The difference between DeiT-S/16 and Swin-Tiny is only 0.0010, and both models therefore provide effectively comparable region-level performance under the present evaluation protocol. Their results suggest that transformer-based representations are well suited to recognizing local symptom patterns within the annotated regions.

The convolutional models nevertheless remain competitive. EfficientNet-B0 and MobileNetV3-Large achieve accuracies of 0.9015 and 0.9010, respectively, exceeding several larger convolutional architectures. ResNet-50 improves over ResNet-18 from 0.8842 to 0.8975, showing a moderate benefit from the deeper residual architecture. However, increasing model scale does not consistently improve performance. EfficientNet-B0 outperforms EfficientNet-B3, while ConvNeXt-Tiny obtains an accuracy of 0.8911 despite its more recent convolutional design. These results indicate that architectural capacity alone does not determine performance on the extracted HERMOS regions.

The HERMOS results should not be compared directly with the complete-image classification results reported in Section~\ref{subsec:classification_results}. The region-level models receive manually localized crops in which the relevant image area has already been identified. At the same time, the regions originate from field-acquired source images and exhibit variation in region dimensions, symptom scale, illumination, background content, and the amount of surrounding leaf tissue. The reported accuracy therefore characterizes discrimination among annotated local regions rather than complete-image disease recognition or automatic disease localization.

\subsection{Object Detection Benchmark Results}
\label{subsec:detection_results}

Table~\ref{tab:detection_results} reports the object-detection results on the final evaluation partition of each dataset. Performance is measured using mAP@50 and mAP@50:95. The first metric evaluates detections at an intersection-over-union threshold of 0.5, whereas the second averages performance over thresholds from 0.5 to 0.95 and therefore places greater
emphasis on precise localization.

\begin{table}[H]
\caption{Object-detection performance on the final evaluation partitions of the Mildew symptom dataset, the LDD leaf-only subset, and FD-Confounders. Results are reported using mAP@50 and mAP@50:95. The best result for each dataset and metric is shown in bold.}
\label{tab:detection_results}
\centering
\small
\begin{tabular}{l l c c}
\toprule
\textbf{Dataset} &
\textbf{Model} &
\textbf{mAP@50} &
\textbf{mAP@50:95} \\
\midrule

Mildew symptom dataset
& YOLOv8n       & 0.7358 & 0.4791 \\
& YOLO11n       & 0.7340 & 0.4822 \\
& YOLO26n       & 0.7363 & 0.4810 \\
& YOLO26s       & 0.7551 & 0.4972 \\
& RF-DETR-Nano  & \textbf{0.7853} & \textbf{0.5209} \\
\midrule

LDD leaf-only subset
& YOLOv8n       & 0.4875 & 0.3082 \\
& YOLO11n       & 0.5238 & 0.3404 \\
& YOLO26n       & 0.5752 & 0.3785 \\
& YOLO26s       & 0.5939 & 0.4160 \\
& RF-DETR-Nano  & \textbf{0.6114} & \textbf{0.4273} \\
\midrule

FD-Confounders
& YOLOv8n       & 0.4958 & 0.2647 \\
& YOLO11n       & 0.5305 & 0.2837 \\
& YOLO26n       & 0.5126 & 0.2714 \\
& YOLO26s       & \textbf{0.5662} & \textbf{0.3109} \\
& RF-DETR-Nano  & 0.5630 & 0.2935 \\
\bottomrule
\end{tabular}
\end{table}

The results show that detection difficulty depends strongly on the dataset and its annotation definition. The highest aggregate performance is obtained on the Mildew symptom dataset, where RF-DETR-Nano achieves an mAP@50 of 0.7853 and an mAP@50:95 of 0.5209. The four YOLO variants also produce relatively similar results on this dataset. Their mAP@50 values range from
0.7340 to 0.7551, while their mAP@50:95 values range from 0.4791 to 0.4972. Among the YOLO models, YOLO26s performs best on both metrics.

The small differences among YOLOv8n, YOLO11n, and YOLO26n on the Mildew symptom dataset indicate that changing the detector generation at the nano scale provides only limited improvement for this particular task. Increasing the capacity from YOLO26n to YOLO26s produces a clearer improvement, raising mAP@50 from 0.7363 to 0.7551 and mAP@50:95 from 0.4810 to 0.4972. RF-DETR-Nano further improves the two metrics by approximately 3.0 and 2.4 percentage points, respectively, over YOLO26s.

The LDD leaf-only subset produces lower detection scores than the Mildew symptom dataset but shows a clearer progression across the YOLO generations. YOLOv8n obtains an mAP@50 of 0.4875 and an mAP@50:95 of 0.3082. YOLO11n improves these values to 0.5238 and 0.3404, while YOLO26n reaches 0.5752 and 0.3785. Increasing the latest architecture from the nano to the small variant provides a further improvement, with YOLO26s reaching 0.5939 mAP@50 and 0.4160 mAP@50:95. RF-DETR-Nano achieves the best LDD results, with corresponding values of 0.6114 and 0.4273.

The progressive improvement from YOLOv8n to YOLO11n and YOLO26n on LDD suggests that the newer feature-extraction and detection designs are beneficial for this multi-class dataset. The additional capacity of YOLO26s is also useful, particularly under the stricter mAP@50:95 metric. Nevertheless, RF-DETR-Nano exceeds YOLO26s by only 0.0175 in mAP@50 and 0.0113 in mAP@50:95. The two models therefore provide comparatively close performance on LDD despite their different detection paradigms.

FD-Confounders is the most difficult dataset under the stricter localization metric. YOLO26s achieves the highest results, with an mAP@50 of 0.5662 and an mAP@50:95 of 0.3109. RF-DETR-Nano obtains a nearly identical mAP@50 of 0.5630 but a lower mAP@50:95 of 0.2935. YOLO11n follows with 0.5305 and 0.2837, respectively. YOLO26n does not improve over YOLO11n on this dataset, indicating that architectural progression at the nano scale does not guarantee consistent gains across annotation settings.

The lower FD-Confounders results are consistent with the characteristics of the dataset. Its images depict complete vineyard scenes containing many overlapping leaves, while the detector must distinguish Flavescence dorée-symptomatic leaves, Esca-symptomatic leaves, and leaves displaying visually confounding symptoms. The task therefore combines dense localization with fine-grained diagnostic discrimination. The relatively large difference between mAP@50 and mAP@50:95 further indicates that obtaining sufficiently precise leaf boundaries is more difficult than producing approximately correct detections.

Across all three datasets, YOLO26s is the strongest YOLO-family model. It outperforms the nano variants on every dataset under both evaluation metrics. RF-DETR-Nano achieves the highest results on the Mildew symptom dataset and LDD, whereas YOLO26s performs best on FD-Confounders. The advantage of the transformer-based detector is therefore dataset-dependent rather than universal.

The comparison also demonstrates that variation among datasets is larger than variation among models evaluated on the same dataset. The best mAP@50:95 ranges from 0.5209 on the Mildew symptom dataset to 0.4273 on LDD and 0.3109 on FD-Confounders. These values should not be interpreted as a direct ranking of dataset quality because the datasets represent different tasks. The Mildew symptom dataset contains directly annotated local symptom regions, LDD contains polygon-derived bounding boxes around local leaf symptoms, and FD-Confounders targets complete diagnostically relevant leaves within complex whole-vine scenes. The results therefore illustrate how annotation granularity, target scale, number of categories, scene complexity, and diagnostic similarity influence detection performance.

\subsection{Cross-Dataset Object Detection Generalization}
\label{subsec:cross_dataset_detection_results}
Cross-dataset object detection was evaluated using YOLO26s as the representative detector. YOLO26s was selected because it achieved the strongest overall performance among the YOLO-family models in the within-dataset benchmark. The same architecture and training protocol were used in both transfer directions, allowing the experiment to focus on the effect of dataset and annotation shift rather than differences between detector architectures.

The evaluation was performed between the Mildew symptom dataset and the LDD leaf-only subset, which are the only two detection datasets in the benchmark that share directly harmonizable disease categories, namely downy mildew and powdery mildew. For each source dataset, YOLO26s was retrained using only these two categories. The resulting model was first evaluated on the corresponding source-domain test partition to establish a two-class within-dataset reference and was then evaluated directly on the held-out test partition of the other dataset. No target-domain images were used for training, model selection, confidence calibration, fine-tuning, or adaptation.

Although both datasets annotate local disease manifestations rather than complete leaves, their annotation procedures are not equivalent. The Mildew symptom dataset uses directly defined bounding boxes around visible downy- and powdery-mildew symptoms, whereas the retained LDD targets are enclosing boxes derived from polygon-level annotations of local leaf symptoms. The datasets also differ in symptom scale, annotation density, image resolution, background composition, acquisition conditions, and class distribution. Consequently, the experiment evaluates transfer under both visual-domain shift and annotation-protocol shift.

Table~\ref{tab:cross_dataset_detection_results} reports both the source-domain reference results and the corresponding cross-dataset results. The LDD test partition contains 77 images and 727 annotations belonging to the two shared classes, whereas the Mildew symptom test partition contains 347 images and 4,840 annotations.

\begin{table}[H]
\centering
\caption{Within- and cross-dataset object-detection performance obtained using YOLO26s after retraining with only the two shared categories, downy mildew and powdery mildew. The within-dataset rows provide source-domain reference performance for the same two-class models used in the transfer experiments. Cross-dataset evaluation was performed directly on the target test partition without target-domain adaptation.}
\label{tab:cross_dataset_detection_results}

\small
\setlength{\tabcolsep}{2.5pt}
\renewcommand{\arraystretch}{1.15}

\begin{adjustbox}{max width=\linewidth}
\begin{tabular}{
    @{}
    >{\raggedright\arraybackslash}p{3.1cm}
    >{\raggedright\arraybackslash}p{3.1cm}
    >{\centering\arraybackslash}m{1.45cm}
    >{\centering\arraybackslash}m{1.35cm}
    >{\centering\arraybackslash}m{1.55cm}
    >{\centering\arraybackslash}m{1.90cm}
    @{}
}
\toprule

\makecell[l]{\textbf{Training}\\\textbf{dataset}} &
\makecell[l]{\textbf{Test}\\\textbf{dataset}} &
\makecell[c]{\textbf{Test}\\\textbf{images}} &
\makecell[c]{\textbf{Box}\\\textbf{count}} &
\makecell[c]{\textbf{mAP}\\\textbf{@0.5}} &
\makecell[c]{\textbf{mAP}\\\textbf{@0.5:0.95}} \\

\midrule

LDD leaf-only subset
& LDD leaf-only subset
& 77
& 727
& 0.3710
& 0.2079 \\

LDD leaf-only subset
& Mildew symptom dataset
& 347
& 4,840
& 0.0010
& 0.0003 \\

\midrule

Mildew symptom dataset
& Mildew symptom dataset
& 347
& 4,840
& 0.7551
& 0.4972 \\

Mildew symptom dataset
& LDD leaf-only subset
& 77
& 727
& 0.0016
& 0.0008 \\

\bottomrule
\end{tabular}
\end{adjustbox}
\end{table}

The source-domain reference results confirm that the two-class YOLO26s models learned meaningful detection functions on their respective training domains before transfer was attempted. The model trained on LDD achieves an mAP@50 of 0.3710 and an mAP@50:95 of 0.2079 when evaluated on the LDD test partition. The model trained on the Mildew symptom dataset performs substantially better within its own domain, reaching 0.7551 and 0.4972, respectively. The difference between these source-domain results is itself consistent with the different annotation characteristics and detection difficulty of the two datasets.

In contrast, direct cross-dataset transfer results in an almost complete performance collapse in both directions. The LDD-trained model obtains only 0.0010 mAP@50 and 0.0003 mAP@50:95 when evaluated on the Mildew symptom dataset. In the reverse direction, the model trained on the Mildew symptom dataset reaches 0.0016 mAP@50 and 0.0008 mAP@50:95 on LDD. Thus, less than 0.3\% of the corresponding source-domain mAP@50 is retained in either transfer direction. The similarly low results in both directions indicate that the generalization problem is not confined to one particular choice of source and target dataset.

Importantly, the source-domain reference rows distinguish cross-dataset failure from unsuccessful two-class retraining. Although the two-class LDD task remains substantially more difficult than the Mildew symptom task, YOLO26s achieves measurable within-dataset detection performance on both sources. The near-zero target-domain results therefore reflect a severe loss of transferability rather than a detector that failed to learn the source task.

The transfer failure also cannot be attributed to incompatible class names, because both models were retrained using exactly the same two-category label space. Instead, nominally identical disease categories appear to define substantially different visual detection problems across the two datasets. In LDD, polygon annotations are converted to enclosing bounding boxes, and the retained targets range from small isolated lesions to larger affected leaf regions. The Mildew symptom dataset uses directly defined local bounding boxes and contains many small, spatially distributed annotations within dense foliage. A detector trained under one annotation convention may therefore learn characteristic target scales, box geometries, symptom boundaries, and contextual cues that are poorly aligned with those of the other dataset.

Class composition provides an additional source of shift. Across the processed Mildew symptom dataset, downy mildew represents 84.3\% of the annotations and powdery mildew 15.7\%. In the corresponding two-class LDD subset, the distribution is considerably more balanced, with 2,292 downy-mildew and 1,933 powdery-mildew annotations. Such differences can affect the learned decision boundary and the confidence distribution of the detector.

The datasets additionally differ in image framing, background complexity, symptom severity, acquisition distance, illumination, grapevine variety, and the amount of contextual leaf tissue surrounding annotated regions. Under the strict zero-adaptation protocol, these differences can affect both class prediction and the spatial agreement between predicted and reference boxes.

For this source--target pair and the evaluated YOLO26s detector, the results show that class-label harmonization alone is insufficient to obtain reliable cross-dataset grape disease detection. Compatibility must also be considered at the level of annotation semantics, target scale, bounding-box construction, acquisition protocol, and symptom appearance. Accordingly, strong within-dataset mAP should not by itself be interpreted as evidence that a detector will generalize to an independently acquired dataset containing diseases with the same names.

Improving transfer between these datasets would require further investigation. Potential directions include annotation harmonization, multi-dataset training, domain adaptation, or fine-tuning using labelled target-domain examples. Joint training would be particularly informative if the two annotation protocols could first be mapped to a more consistent definition of a detection target. The present zero-adaptation experiment nevertheless provides a useful stress test: despite an identical two-class taxonomy and measurable source-domain performance, direct transfer between the two datasets remains extremely limited.

\section{Conclusions}
\label{sec:conclusion}

This study presented a dataset-centric benchmark of deep learning methods for grape leaf disease classification and detection. Rather than treating datasets only as interchangeable sources of training and test samples, the benchmark examined how acquisition conditions, disease taxonomy, annotation granularity, class distribution, image provenance, and scene complexity influence model performance. The study considered three complementary evaluation settings: image-level classification, region-level classification, and object-level detection. This separation was necessary because these settings differ in their input units, localization assumptions, annotation requirements, and practical interpretation.

The image-level classification results showed that several controlled, augmented, or derivative datasets are close to saturation under conventional within-dataset evaluation. PlantVillage, PDR2018, the Grape Leaf Disease Dataset, the Grape Leaf Disease Augmented Dataset, and the New Plant Diseases grape subset produced near-perfect or perfect accuracy for most evaluated architectures. Under these conditions, differences between modern convolutional and transformer-based models were generally small, indicating that high within-dataset accuracy does not necessarily provide a sufficiently discriminative assessment of model robustness. GL-Portugal also produced consistently high accuracy despite being acquired in vineyard environments, although its predominantly visible, leaf-centered subjects represent less complex conditions than unconstrained canopy-level monitoring.

In contrast, GVLiD and GLHCD were more challenging for different reasons. GVLiD introduces greater field variability and scene complexity, whereas GLHCD combines a broader five-class taxonomy with greater visual heterogeneity among several categories. The analysis of dataset provenance additionally identified exact image overlap among several controlled or derivative datasets, showing that apparently strong cross-dataset transfer within this group must be interpreted cautiously. More importantly, transfer between GVLiD and the controlled or derivative datasets resulted in substantial accuracy degradation. These findings demonstrate that compatible class names and high within-dataset performance do not ensure generalization to an independently acquired field domain.

The HERMOS experiment provided a complementary region-level evaluation in which annotated regions were extracted before classification. DeiT-S/16 achieved the highest accuracy of 0.9148, followed closely by Swin-Tiny with 0.9138. However, the relatively narrow performance range across the evaluated architectures showed that several convolutional and transformer-based models were capable of learning useful representations from the localized regions. These results characterize discrimination after the relevant image area has already been identified and should therefore not be compared directly with complete-image classification accuracy or object-detection performance.

A separate Grad-CAM analysis was performed using a correctly classified GL-Portugal test image. The representative ResNet-50 prediction assigned a probability of 92.9\% to the downy-mildew class, while the activation map was concentrated primarily on the visibly affected region of the foreground leaf. This result provides a qualitative indication that the classifier used disease-relevant visual evidence in this example. However, the Grad-CAM map was included only as an illustrative interpretation of model behaviour and should not be considered a precise symptom segmentation or a basis for model ranking.

The object-detection benchmark showed that performance was influenced more strongly by the dataset and its annotation definition than by detector generation alone. RF-DETR-Nano achieved the highest performance on the Mildew symptom dataset and the LDD leaf-only subset, reaching mAP@50:95 values of 0.5209 and 0.4273, respectively. YOLO26s was the strongest YOLO-family detector across all three datasets and achieved the best result on FD-Confounders, with an mAP@50:95 of 0.3109. No detector was universally superior across all annotation settings. In particular, FD-Confounders remained the most demanding dataset because it combines dense complete-leaf localization, overlapping vegetation, and fine-grained discrimination among Flavescence dorée, Esca, and visually confounding symptoms.

The YOLO26s cross-dataset detection experiment provided particularly clear evidence of dataset dependence. Although the Mildew symptom dataset and LDD share the categories downy mildew and powdery mildew, direct transfer between them produced near-zero performance. Training on LDD and testing on the Mildew symptom dataset resulted in an mAP@50:95 of 0.0003, while the reverse direction produced 0.0008. The corresponding source-domain values were 0.2079 and 0.4972, respectively, confirming that the collapse reflects loss of cross-dataset transferability rather than unsuccessful source-domain training. These results show, for this source--target pair and detector, that semantic label correspondence alone is insufficient to define a transferable detection task.

Several limitations should be considered when interpreting the results. First, direct comparison across detection datasets remains constrained by differences in annotation units, since the evaluated targets include local symptom regions, polygon-derived disease boxes, and complete diagnostically relevant leaves. Second, only datasets with sufficiently compatible taxonomies could be included in cross-dataset experiments. Third, some classification datasets contain augmented, derivative, or overlapping content, limiting their value as fully independent evaluation domains. Fourth, the HERMOS annotations are not sufficiently exhaustive for conventional full-image detection evaluation and were therefore used only for region-level classification. Finally, the benchmark evaluates the available public datasets and does not establish clinical or agronomic validity under all cultivars, disease stages, geographic regions, acquisition devices, and growing seasons.

Future work should prioritize the construction of larger and more diverse field-acquired datasets with expert-verified labels, exhaustive spatial annotations, and clearly documented acquisition protocols. Greater standardization of disease terminology and annotation semantics would make cross-dataset comparison more reliable. Multi-dataset training, domain adaptation, annotation harmonization, self-supervised pretraining, and target-domain fine-tuning should be investigated as possible strategies for improving generalization. Additional research could also extend the benchmark toward instance segmentation, symptom-severity estimation, uncertainty calibration, video-based temporal analysis, and evaluation across cultivars, vineyards, seasons, and imaging devices.

Overall, the results demonstrate that reliable grape leaf disease recognition depends not only on selecting an accurate architecture, but also on the quality, diversity, provenance, and annotation design of the data used for training and evaluation. High performance on a single controlled or visually standardized dataset should not be interpreted as sufficient evidence of readiness for practical vineyard deployment. Robust progress will require evaluation protocols that combine realistic field data, transparent dataset analysis, compatible annotation definitions, and cross-dataset validation.

\bibliographystyle{elsarticle-num} 
\bibliography{references}

\end{document}